\documentclass[12pt,notitlepage]{article}
\usepackage{graphicx}

\usepackage{ amsmath, amscd, verbatim, amssymb }
\usepackage[round]{natbib}
\usepackage{algorithm}
\usepackage{algorithmic}

\DeclareMathOperator*{\argmin}{arg\,min}
\DeclareMathOperator*{\argmax}{arg\,max}

\usepackage{anysize}

\marginsize{3 cm}{2 cm}{1 cm} {2 cm}
\begin{document}

\author{Roberto Aldave\\
\small{\texttt{Roberto.Aldave@USherbrooke.ca}} \\
				Jean-Pierre Dussault\\
\small{\texttt{Jean-Pierre.Dussault@USherbrooke.ca}}	\\
\\
				D\'epartement d'Informatique\\
				Universit\'e de Sherbrooke\\
				Sherbrooke, Qu\'ebec, J1K 2R1, CANADA}
				
\title{ Systematic Ensemble Learning for Regression}
\date{\vspace{-5ex}}
\maketitle

\begin{abstract}
The motivation of this work is to improve the performance of standard stacking approaches or ensembles, which are composed of simple, heterogeneous base models, through the integration of the generation and selection stages for regression problems. We propose two extensions to the standard stacking approach. In the first extension we combine a set of  standard stacking approaches into an ensemble of ensembles using a two-step ensemble learning in the regression setting. The second extension consists of two parts. In the initial part a diversity mechanism is injected into the original training data set, systematically generating different training subsets or partitions, and corresponding ensembles of ensembles. In the final part after measuring the quality of the different partitions or ensembles, a max-min rule-based selection algorithm is 
used to select the most appropriate ensemble/partition on which to make the final prediction. We show, based on experiments over a broad range of data sets, that the second extension performs better than the best of the standard stacking approaches, and is as good as the oracle of databases, which has the best base model selected by cross-validation  for each data set. In addition to that, the second extension performs better than two state-of-the-art ensemble methods for regression, and it is as good as a third state-of-the-art ensemble method.

\end{abstract}

\section{Introduction}
\label{intro}
Recent developments in the machine learning community have brought innovative ideas and methods to solve problems in data mining and predictive analytics with important applications in the business world such as credit risk, fraud detection, survey direct response, customer segmentation, etc.. Among such methodologies ensemble learning has been an active research area within the last two decades, where it is believed that combining the predictions of different models or base models, e.g., by weighted averaging, can bring an improvement in the prediction error in comparison to the error provided by the best individual model or base model. Alternative ways that we will be using throughout the paper to refer to base models are base learner, level-0 learner or level-0 models. 

The setting of the approximation for regression we are considering is in the context of supervised learning: Given an historical dataset $U$ with $N$ training samples $(\boldsymbol{x}_n,y_n)$, $n=1, ...,N$, where $y=f^*(\boldsymbol{x})+\epsilon$, and $\epsilon$ is a stochastic error, the task is to approximate the unknown function $f^*$ by $f$, so we fit a model $f(\boldsymbol{x},\boldsymbol{\theta})$ to the training data by minimizing the square error

\begin{equation}
  \hat{\boldsymbol{\theta}} = \argmin_{\boldsymbol{\theta}} \sum (y_n - f(\boldsymbol{x}_n,\boldsymbol{\theta}) )^2 
 \end{equation}
 
Interested readers can consult \citet{gentle02} and \citet{berthold99} for a deeper discussion 
about this topic. The aim of this work is to approximate the unknown function $f^*(\boldsymbol{x})$ using a linear combination of base learners, or learning algorithms. So, instead of a single estimator $f$ we have a collection of them; $f_1, f_2, ..., f_J$. We then learn each individual $\hat{f}_j(\boldsymbol{x}) =f_j(\boldsymbol{x},\hat{\boldsymbol{\theta}})$ model separately using the error function in Equation (1) on the training samples. Once this is achieved, the outputs of the learned models are combined  resulting in an approximating ensemble function  $\tilde{f}(\boldsymbol{x})$:

\[
\tilde{f}(\boldsymbol{x},\boldsymbol{\alpha}) = \sum_{j} \alpha_{j} \hat{f}_{j}(\boldsymbol{x}) 
\]

The above function may be used to make predictions of $y_0=f^*(\boldsymbol{x}_0)+\epsilon$ for a new instance $\boldsymbol{x_0}$. There are different ways to determine the $\boldsymbol{\alpha}$ weight vector. In Section 2.3 we introduce the way we are going to calculate it in this paper.

The expected error of $\tilde{f}(\boldsymbol{x}_0)$, or $Mean$ $Square$ $Error$ $(MSE)$, is defined as $MSE(\tilde{f}(\boldsymbol{x}_0))\allowbreak = E[(y_0 - \tilde{f}(\boldsymbol{x}_0))^2]$. The $MSE$ can be used to measure the generalization error of a predictor on new instances, and can be decomposed as the sum of two prediction errors \citep{yu06bias}, the variance, $Var(\tilde{f}(\boldsymbol{x}_0))$, or spread of $\tilde{f}(\boldsymbol{x}_0)$ around its average prediction, $\bar{f}(\boldsymbol{x})=E[\tilde{f}(\boldsymbol{x}_0)]$, and the square of the bias, $Bias^2(\tilde{f}(\boldsymbol{x}_0))$, or amount by which the average prediction of $\tilde{f}(\boldsymbol{x}_0)$ differs from $f^*(\boldsymbol{x}_0)$.

$MSE{( \tilde{f}(\boldsymbol{x}_0) )} = Var(\tilde{f}(\boldsymbol{x}_0)) + Bias^2(\tilde{f}(\boldsymbol{x}_0)) = E[\tilde{f}(\boldsymbol{x}_0) - \bar{f}(\boldsymbol{x}_0) ] + [\bar{f}(\boldsymbol{x}_0) - f^*(\boldsymbol{x}_0) ]^2 $

This is known as the bias-variance dilemma. The variance term can be further decomposed for an ensemble as follows \citep{yu06bias} :
\begin{equation}
\begin{split}
Var(\tilde{f}(\boldsymbol{x}_0)) = E[\tilde{f}(\boldsymbol{x}_0) - \bar{f}(\boldsymbol{x}_0) ] =  
 E[{(\tilde{f}(\boldsymbol{x}_0)-E(\tilde{f}(\boldsymbol{x}_0)) )}^2 ] = \\ \sum_{j=1}^{J}\alpha_j^2(E[f_j^2(\boldsymbol{x}_0)]-E^2[f_j(\boldsymbol{x}_0)]) + \\
 2\sum_{j<k}\alpha_j\alpha_k(E[f_j(\boldsymbol{x}_0)f_k(\boldsymbol{x}_0)] - E[f_j(\boldsymbol{x}_0)]E[f_k(\boldsymbol{x}_0)] )
\end{split}
\end{equation}

The expectation operator is taken with respect to the historical, or training dataset $U$. The first sum contains the lower limit of the ensemble variance, which is the weighted mean of the variance of ensemble members. The second sum contains the cross terms of the ensemble members, which disappears if the models are completely uncorrelated \citep{krogh1997statistical}.

According to the principle of the above bias-variance trade-off, an ensemble consisting of diverse models with much disagreement is more likely to have a good generalization performance \citep{yu06bias}. So, how to generate diverse models is very important. For example, for neural networks, \citet{keung06} suggest to initialize different weights for each neural network models, training neural networks with different training subsets, etc.. In our work, the main driver to achieve that purpose will be to inject a diversity mechanism into the cross-validation procedure, which will generate new partitions on which corresponding ensembles will be trained. We will then select a partition, or ensemble where the correlation between their base components satisfies a minimum correlation criterion, which is intended to have an effect in second sum of Equation (2) to lower the variance.  

The aim is to establish a trade-off between bias and variance when training any predictor or classifier with good generalization, i.e., with good  performance when applying it on new instances, not present in the training mechanism. It is generally accepted that combining different classifiers can reduce the variance while not affecting the bias \citep{bishop95}, \citep{hashem97}. It should be mentioned briefly that another approach for improving the generalization error is based on statistical learning theory \citep{vapnick98}, which has its origins with \citet{valiant84} in $probably$ $approximately$ $correct$ or PAC learning. The goal of PAC learning is to understand how large a dataset needs to be in order to give good generalization, we will not use this approach in our work. 

According to \citet{mendes12}, most of the research focuses on one specific approach to build the ensemble, e.g., sub-sampling from the training set or manipulating the induction algorithm, and further investigation is necessary to achieve gains by combining several approaches. This observation of using different ideas along with the potential gain we can achieve in prediction accuracy by a joint design of ensemble generation, and integration steps, have motivated us to investigate the performance of standard stacking approaches, and develop new extensions to stacking for regression.

In this work we present methodology that we have developed within the framework of ensemble learning through stacked generalization for regression.  The main idea behind stacking is to generate a meta-training set from the model predictions on validation sets with the aim to integrate them through a learning mechanism, e.g. by weighted average. 

We propose to use a set of simple base learners, and consider the selection of the best base learner as a benchmark to compare the performance of the stacking approaches. Our empirical evaluation shows that the oracle composed of best base learners performs better than the best of the standard stacking approaches used in this work. In this paper we will refer to any of the standard stacking approaches as level-1 learner, level-1 ensemble, or base ensemble as well.

We propose a couple of extensions to the standard stacking approach. The first extension considers combining standard stackings into an ensemble of level-1 learners, using a two-step level-2 ensemble learning, i.e., we are making the ensemble grow. The two-step level-2 ensemble learning scheme introduced here provides a way to measure the diversity of the base ensembles, or level-1 learners through computing pairwise correlations between its components, which will be exploited in a second extension with the aim to improve the prediction accuracy of the two-step level-2 ensemble. 

The second extension to standard stacking is composed of two parts. The initial part consists of an extra diversity mechanism injected in the data generation which creates different cross-validation partitions from current cross-validation partition, on which new level-2 ensembles are trained. The final part, it is an algorithm based on computed level-1 learner correlation criterion, and ranking based-criterion, applied after two-step level-2 ensemble learning, which selects the most appropriate ensemble of standard stacking approaches, and/or corresponding data partition on which to make the final prediction. We show in this work that the latter extension performs better than the best of the standard stacking approaches, or level-1 ensembles, and it is as good as the best base model  by cross-validation. We also show that such extension performs better than two of state-of-the-art ensemble learning algorithms, and it is as good as a third state-of-the-art method.

In Section 2, we summarize the state-of-the-art, the stacking framework, and survey some recent approaches of stacking regression.  

In Section 3, we introduce a set of base models, and level-1 learning which consists of standard stacking approaches, each composed of different combinations of base models.

In Section 4, we present our first extension to standard stacking, level-2 learning, a mixture of ensembles of level-1 learners. We introduce two different learning schemes we have developed: two-step ensemble learning, and all-at-once ensemble learning. The former has much richer structure than the latter, which will be exploited in the algorithm presented later in Section 5.

In Section 5, we introduce the second extension to stacking, which initially consists of building a diversity mechanism to generate different $CV$ partitions by perturbing the current $CV$ partition; and finally a partition selection is done, using either a computed level-1 learner correlation-based criterion, or a ranking-based criterion within a max-min rule-based heuristic algorithm. At the end, we present such algorithm for systematic ensemble learning based on generating the partitions in an orderly way.

In Section 6, the experimental setup is described for the comparison of the different ensemble approaches, the oracle of databases, and three state of the art ensemble learning algorithms. Section 7 presents and discusses the experimental results, and Section 8 concludes. Our results match in error performance the ones obtained with the oracle of databases. In addition to that our results performed better than two of the state-of-the-art ensemble algorithms for regression, and it also matches the ones obtained with a third state-of-the-art ensemble method for regression.\\

\section{State-of-the-Art}
\label{sec:2}
We review some background information about ensemble learning in Section 2.1, followed by the framework of cross-validation in Section 2.2, and in Section 2.3, we present stacking introduced by \citet{wolpert92}. We then describe in Section 2.4 recent stacking approaches of incremental learning, model integration, and we review as well some recent studies for regression, less referred in literature. 

\subsection{Ensemble learning}
\label{sec:2.1}
Typically, ensemble learning consists of three phases, the first is ensemble generation which generates a set of base models. Second is ensemble pruning where the ensemble is pruned by eliminating some of the base models generated earlier. Finally, in the integration step a strategy is used to obtain the prediction for new cases.

Most of the work done in ensemble generation is about generating a single learning algorithm known as homogeneous learning. To ensure a level of diversity amongst the base models, one of the following strategies are used \citep{rooney07, dietterich00}: Varying the learning parameters of the learning algorithm, Varying the training data employed such as Bagging \citep{breiman96b}, and Boosting \citep{drucker97}, both use re-sampling techniques to obtain different training sets for each of the classifiers. Varying the feature set employed, such as Random Subspace \citep{ho98}, which is a feature selection algorithm, where each base learner corresponds to a different subset of attributes from the feature space. In Randomize Outputs approach, rather than present different regressors with different samples of input data, each regressor is presented with the same training data, but with output values for each instance perturbed by a randomization process \citep{breiman00}, \citep{chri03}.

Another approach, known as heterogeneous learning, is to generate models using different learning algorithms where each model is different from each other. The conventional approach where each base model has been built with a different learning algorithm but the same training data is called $strict$, whereas a $non$-$strict$ heterogeneous ensemble set  where the base models are built from more than one learning algorithm, but each base model is not required to be built with a distinct learning algorithm \citep{rooney07}. The base models in this last approach can be generated from homogeneous and/or heterogeneous learning.

In pruning step the objective is to reduce the size of the ensemble with the aim to improve the prediction accuracy \citep{mendes12}. It has been shown that given a set of $N$ base models it is possible that an ensemble performs better if only uses a subset of such models instead of all of them. \citet{zhou02} reduces the size using a measure of diversity and accuracy, where the  diversity is measured with the positive correlation of base models in their prediction errors. They used homogeneous learning for stacked regression. 

The integration approach can be performed in two ways, either by combining the predictions of different models, or selecting one of the base models to make the final prediction \citep{rooney07}. Integration can be classified as $static$ or $dynamic$. It is static when for any new instance the combination of predictions is done globally in the same way, based on the whole instance set; whereas for different new instances the dynamic approach combines the predictions based on different localized regions of the instance space. For the  static case the simplest way to combine is to average the predictions of the base models, which is referred to as the unweighted averaging \citep{perrone93}. Among the weighted averaging techniques we have one based on the performance of individual ensemble members \citep{merz99}, or in the case of Artificial Neural Networks $(ANN)$ to penalize correlations \citep{rosen96}, and \citep{liu99} where the base models produced are negatively correlated. The simplest static selection approach is to choose the model with minimum cross-validation overall training error referred to as Selection by Cross-Validation \citep{schaffer93}, o also as Select Best \citep{dzeroski04}. It is very common to use these methods to benchmark ensemble approaches. 

\subsection{Cross-validation}
\label{sec:2.2}
Cross-validation $(CV)$ is a resampling technique often used in data mining for model selection and estimation of the prediction error in regression and classification problems. In order to measure the performance of one model the historical set $U$ is split in two parts, the training $S$ and the validation $V$. The training part $S$ is used to fit a model which is then used to predict the data on the validation set $V$. Because we know in advance the actual values for the predicted variable in the validation set $V$, then a performance measure such as the $MSE$, defined below lines for cross-validation, is used to estimate the accuracy achieved by the model. 

We should point out that estimating the prediction error must be done on data points on $V$ which are outside of the data $S$ that was used for model estimation to prevent from having \emph{overfitting}. Overfitting the training data happens when the model fits so well the training data that results in poor generalization when it is used to predict new instances not present in the fitting phase.

The general $K$-fold cross-validation procedure is as follows:

1. Randomly split the data in $K$ almost equal parts $L_1,L_2,...,L_K$.

2. For each $k$th fold define $L_k$ and $L^{(-k)}=L-L_k$ to be the validation and training sets for the $k$th fold of a $K$-fold cross-validation.

3. Invoke the learning algorithm on the training set $L^{(-k)}$ to estimate a model $\hat{f}^{(-k)}$
and calculate its prediction error on the $L_k$ fold.

4. Repeat the above procedure for $k=1,2,...,K$.

Let $\boldsymbol{x}_n $ be the $n$-th observation in $L_k$ and be $\hat{f}^{-k}(\boldsymbol{x}_n)$ the prediction of the estimated model on $\boldsymbol{x}_n $.

The cross-validation estimate of mean square error is:

\[
  \widehat{MSE}_{CV_k}=\widehat{MSE}_{CV}(\hat{f}^{(-k)}) = \frac{1}{\#(L_k)} \sum_{n \in L_k}(y_n - \hat{f}^{(-k)}(\boldsymbol{x}_n))^2
\]
We will be using the cross-validated mean square error normalized by the sample variance for model performance, which is defined as 

\[
  \widehat{NMSE}_{CV_k}= \frac{1}{\sum_{n \in L_k}(y_n - \bar{y})^2} \sum_{n \in L_k}(y_n - \hat{f}^{(-k)}(\boldsymbol{x}_n))^2
\]

\subsection{Stacking}
\label{sec:2.3}
Stacking is the abbreviation used to refer to stacked generalization \citep{wolpert92}, 
it has been used for regression tasks \citep{breiman96}, and classification purposes \citep{leblanc93}. Further development has been done by \citet{ting99}. For more recent extensions of stacking see \citet{todorovski03}.
In stacking the general idea is to have level-0 models which consist of a set of $J$ different predictors whose task is to generate predictions based on different partitions of a dataset. Those $J$ level-0 models are the learned models on a training dataset $L$ corresponding to $J$ learning algorithms. 

Given a dataset  $L=\{(y_n,\boldsymbol{x}_n ),n=1,...,N\}$, where $y_n$ is the actual value and $\boldsymbol{x}_n $ is a vector representing the attribute values of the $n$th instance, randomly split the data into $K$ almost equal parts to get $L_k$ and $L^{(-k)}$ the validation and training sets for the $k$th fold of a $K$-fold cross-validation. These datasets constitute the $level-0$ $data$. It should be noted that the validation subsets $L_k$, $k=1,...,K$ are mutually exclusive and that their union gives the whole dataset $L$.

For each instance $\boldsymbol{x}_n $ in $L_k$, the validation set for the $k$th cross-validation fold, let $\hat{f}_{j}^{(-k)}(\boldsymbol{x})=f_{j}^{(-k)}(\boldsymbol{x},\hat{\boldsymbol{\theta}})$ denote $j$-th model fitted on training subset $L^{(-k)}$ and $z_{jn}=\hat{f}_{j}^{(-k)}(\boldsymbol{x}_n)$  denote the prediction of the estimated model $\hat{f}_{j}^{(-k)}$ on $\boldsymbol{x}_n $. At the end of the $K$-fold $CV$ procedure, the data set assembled from the outputs of the $J$ models along the $K$ validation subsets is

\[L_{CV}=\{(y_n,z_{1n},z_{2n},...,z_{Jn}),n=1,...,N\}. \]

The $level-1$ $data$ consists of $ \{(y_n,z_n),n=1,...,N \}$  where \\ $z_n=(z_{1n},z_{2n},...,z_{Jn})$. Then, using some learning algorithm that we call the $level-1$ $generalizer$ to derive from these level-1 data a model $f$ for $y$ for prediction of new data $\boldsymbol{x}$. The $level-1$ $model$ is  

\[ \tilde{f}(\boldsymbol{x},\boldsymbol{\alpha}) = \sum_{j} \alpha_{j}f_{j}(\boldsymbol{x},\hat{\boldsymbol{\theta}}) \] 

The $\boldsymbol{\alpha}$ vector is chosen by minimizing

\[
 \min_{\boldsymbol{\alpha}} \sum_{k} \sum_{(y_n,\boldsymbol{x}_n )\epsilon L_k} (y_n - \sum_{j} \alpha_{j}f^{(-k)}_{j} (\boldsymbol{x}_n),\hat{\boldsymbol{\theta}})^2. 
\]

This prediction is expected to be more accurate than, e.g. the one we get with the best level-1 model selected from cross-validation. 

\citet{breiman96} used as level-0 learners either linear regression with different number of variables, or different collections of shrinkage parameters for ridge regression. Then, he used the level-0 models' outputs, level-1 data, as  input for the level-1 generalizer to minimize the least squares criterion under the constraints $\alpha_j \geq 0$, $j=1,...,J$, with the additional constraint $\sum_j \alpha_{j} = 1$. Breiman concluded that both were better than the corresponding single best model as selected by cross-validation. Stacking improved subset selection more than it improved ridge regression though, which Breiman suggested was due to a greater instability of subset selection. He also stacked together subset and ridge regressions, which results in a predictor more accurate than either ridge, or variable selection.

\subsection{Recent approaches in stacking }
\label{sec:2.4}
The key enabling concept in all ensemble based systems is diversity \citep{polikar07}. Little research, however, has been devoted to constructing ensembles with different base classifiers \citep{zhu10}; heterogeneous ensembles are obtained when more than one learning algorithm is used. In \citet{dzeroski04}, they use heterogeneous base classifiers as base learners, and propose two extensions for stacking in classification problems.  One extension is based on an extended set of  meta-level features, and the other uses multi-response model trees to learn the meta-level. They showed, that the latter performs better than existing methods of stacking approaches, and selecting the best classifier from the ensemble by cross-validation.

In heterogeneous learning the generated base models are expected to be diverse. However, the problem is the lack of control on the diversity of the ensemble components during the generation phase \citep{mendes12}. One  approach, commonly followed in the literature, combines two approaches: using different induction algorithms, e.g., heterogeneous models, mixed with the use of different parameter sets, e.g. different neighbourhood sizes, see \citet{rooney04} and \citet{merz96}. 
Another way to overcome the lack of control in diversity is through ensemble pruning; \citet{caruana04} embedded the ensemble integration phase in the ensemble selection one. \citet{rooney06} extended the technique of stacked regression to prune an ensemble of base learners by taking into account the accuracy, and diversity. The diversity is measured by the positive correlation of base learners in their prediction errors. The above examples show that an important part of the problems at the integration phase can be solved by a joint design of the generation, pruning, when appropriate, and integration phases \citep{mendes12}. 

In this work we propose a set of base learners diverse and accurate, and combine them using stacking into what we already called, standard stacking approach, or level-1 learning. We generate different standard stackings based on different combination of base models. We also combine such standard stackings into an ensemble of level-1 learners through a level-2 ensemble learning giving an extra level of smoothing in the predictions, which is a way to control the diversity of such ensembles. 

There has been considerably less attention given to the area of homogeneity in the area of stacked regression problems \citep{rooney07}.  For example, among the best state-of-the art stacking for regression methods is wMetaComb \citep{rooney07b}, which integrates two different integration methods using homogeneous learning. The method uses a weighted average to combine stacked regression, and a dynamic stacking method, $(DWS)$; the weights are determined based on the error performance of both methods. Tests conducted on 30 regression data sets resulted, on one side, that wMetaComb never looses against stacked regression, where it wins 5, and draws the remaining 25; on the other side, it looses 3 against $DWS$, where it wins 13 and draws 14. A state of the art method using heterogeneous learning for stacking regression is Cocktail ensemble learning \citep{yu07}. They used an error-ambiguity decomposition to analyse the optimal linear combination of two ensembles, and extended it to multiple ensembles via pairwise combinations. They claim that such hybrid ensemble is superior to the base ensemble, simple averaging, and selecting the best ensemble.

\section{Level-1 Ensemble Learning}
\label{sec:3}
It should be noted that our initial work based on standard stacking differs from the literature in two aspects. First, we used a particular set of level-0 models, that we will introduce in Section 3.1, and second, in Section 3.2 we introduce three standard stacking approaches each of them composed of a different combination of base learners. 

\subsection{Level-0 Learners}
\label{sec:3.1}
The $J=4$ learning algorithms we consider in this work are three versions of linear regression, and a fourth learning algorithm which can be seen as a type of non-linear regression. Let $f_1, f_2, ..., f_J$  be the set of $J$ learned base learners of corresponding $J$ different learning algorithms trained on a dataset $L$. Next, we will define the prediction models that we use as first level-0 learners. It is worth to mention that the methodology developed in this work can work with any number $J$ of different base learners. 

For the learned level-0 learner of Constant Regression $(CR)$, or naive regression, we just consider the simple average value of the dependent variable in the training set as the predictor, i.e., $CR=f_1(\boldsymbol{x},\boldsymbol{\theta})=\theta_0=\frac{1}{n}\sum{\boldsymbol{x}_n }$. Second, when we formulate a learned Linear Regression model $(LR)$ on a training dataset, we have in total $I + 1$ terms, corresponding to  $I$ single terms plus the constant term, i.e., $LR=f_2(\boldsymbol{x},\boldsymbol{\theta)}=\theta_0 + \sum_{i=1}^{I}x_i\theta_i=\theta_0 + \boldsymbol{\theta_1}^t \boldsymbol{x}$, , where $\boldsymbol{x}$ and $\boldsymbol{\theta_1}$ are both vectors. Third, we formulate a learned Quadratic Polynomial Regression model $(QR)$ on a training dataset, where we include $I$ single terms corresponding to $I$ independent variables, $I$ quadratic terms corresponding to the squared independent variables, as well as $I(I-1)/2$ cross terms plus one term for the constant term. Let $QR=f_3(\boldsymbol{x},\boldsymbol{\theta})=\theta_0 + \boldsymbol{\theta}_1^t \boldsymbol{x} + \frac{1}{2}\boldsymbol{x}^t \boldsymbol{\Theta}_3 \boldsymbol{x} $ be the matrix representation of the quadratic polynomial, where $\boldsymbol{x}$ and $\boldsymbol{\theta}_1$ are both vectors, and $\boldsymbol{\Theta}_3$ is a symmetric matrix. We are going to introduce a fourth learned level-0 learner trained on a training dataset known as Radial Basis Function Networks $(RBF)$. In next section, we will denote $f_1$ same as $CR$, i.e. $ f_1 \leftarrow CR $. In the same way we define $f_2$ as $f_2 \leftarrow LR$, $f_3$  as $f_3 \leftarrow QR$, and $f_4$ as $f_4 \leftarrow RBF$.  

The radial basis function approach is a very active research discipline particularly in neural networks and penalized splines \citep{ruppert03}. We will be considering radial basis function networks as one type of radial basis model, which can be considered as an statistical model for non-parametric regression, to be introduced shortly.

\vspace{0.5 cm}
\emph{Radial Basis Function Networks}

Given the set of $N$ training points $Z=(y_n,\boldsymbol{x}_n )$ where $n=1,..,N$ and $dim(\boldsymbol{x}_n)=I$. In order to obtain a radial basis function approximation of the set $Z$, a model $f$ of the form \[ f(\boldsymbol{x},\boldsymbol{\theta})= \theta_0 + \sum_{n=1}^{N} \theta_n\phi_n(||\boldsymbol{x},\boldsymbol{x}_n ||) = \sum_{n=0}^{N} \theta_n\phi_n(||\boldsymbol{x},\boldsymbol{x}_n ||)  \]

\noindent where $\phi_n(||\boldsymbol{x},\boldsymbol{x}_n ||) = \phi_n(r_n)$ and, $r_n$ is the radial distance between $x$ and $\boldsymbol{x}_n $, is fitted to the data by minimizing the error function in Equation (1).

In the above model the points $\boldsymbol{x}_n $ are considered as the centers of the radial basis functions. In real applications to localize the radial basis functions on the input vector $\boldsymbol{x}_n $ can be very expensive, so usually a reduced number of centers $M$ where $M < N$ are estimated first. In our work an optimal number of centers is calculated by splitting the historical data in training and validation parts. This will give us the number of radial basis functions of the model. Other parameter that is estimated in this part is the spread parameter $\sigma$ which play a role in the gaussian and multi-quadratic models. We use $k$-means clustering for estimation of centers and $\sigma$'s. Once we estimate these set of parameters, we will estimate the weights $w$'s assuming a fixed number of radial basis functions, and a constant spread for each function. In this way the estimation of weights is linear and can be estimated by standard least squares optimization of the error function as in Equation (1). This is an advantage of this kind of models \citep{bishop95} where we are avoiding to optimize together all these parameters, which would require a non-linear optimization of the weights  where the error function is non-convex, and with many local minima \citep{hastie01}. Another advantage to do it this way is that the interpretation of the model, from the statistical point of view, is very simple as a linear combination of non-linear transformation of the input variables. Nevertheless, this two parameter learning approach is a very active research subject.

There are six basis functions, which are recognized as having useful properties for $RBF$ models \citep{bishop95}: Inverse Multi-quadratic, Gaussian, Multi-quadratic, Thin Plate Spline, Cubic and Linear. For example, the Gaussian is defined as 
$\phi(r) = \exp{-\frac{r^2}{2\sigma^2}}$ for $\sigma>0$ and $r \in {\mathbb R}$, and is considered a local function in the sense that $\phi \rightarrow 0$ as $|r| \rightarrow \infty$. The same property is exhibited by Inverse Multi-quadratic function. On the other hand, the last four are global radial basis functions, which all have the property $\phi \rightarrow \infty$ as $|r| \rightarrow \infty$, e.g. this is valid for Thin Plate Spline, which is defined as $\phi(r)=r^2ln(r)$. For formulations of the other functions please see \citet{bishop95}.

Among the group of local radial basis functions, we are interested on gaussian radial basis function as the main replacement of the quadratic regression model because, we think, it will provide a better non-linear behaviour than the quadratic regression model.

\subsection{Level-1 Learning }
\label{sec:3.2}
We introduce and define in Section 3.2.1, the standard stacking approaches proposed in this work based on different combinations of level-0 learners. 

\subsubsection{Level-1 Learning}

The basic methodology is described next:

\begin{enumerate}

\item We estimate first learned constant regression $f_1$, linear regression $f_2$, and quadratic regression $f_3$ on the training sets of a $K$-fold $CV$. The first one is just the average value taken on the dependent variable values. The last two are calculated from ridge regression by minimizing the penalized sum of squares using Tikhonov regularization. It should be mention that the we used a fixed value for the regularization parameter in the ridge regression estimate, we do not optimize such regularization parameter using cross-validation. We then compute the predictions of each based model on the validation folds of a  $K$-fold $CV$.

Then, we define a level-1 $generalizer$, or learning ensemble $\tilde{f}$ as a linear combination of base learners in set $\{f_1,f_2,f_3\}$: $\tilde{f}(\boldsymbol{x})=\alpha_1f_1(\boldsymbol{x}) + \alpha_2f_2(\boldsymbol{x}) + \alpha_3f_3(\boldsymbol{x})$
where we estimate the weight vector $\boldsymbol{\alpha}$ by $K$-fold cross-validation $(CV)$ based on a stacked generalization procedure to combine the individual model predictions 

For the determination of the weights, we followed what we found in literature, which can be made by optimizing last equation in section 2.3, which we reproduce here for convenience.

\item Optimal Weight Learning from $CV$.

\begin{itemize}

\item Given a dataset  $L=\{(y_n,\boldsymbol{x}_n ),n=1,...,N\}$ do $K$-fold $CV$. 

\begin{equation}
\min_{\boldsymbol{\alpha}} \sum_{k} \sum_{(y_n,\boldsymbol{x}_n )\epsilon L_k} (y_n - \sum_{j} \alpha_{j}f^{(-k)}_{j} (\boldsymbol{x}_n,\hat{\boldsymbol{\theta}}))^2. 
\end{equation}

\[\text{s.t. } \sum_j{\alpha_j}=1.\]

\end{itemize}

In the above equation, the $level-1$ $data$ consists of $ \{(y_n,z_n),n=1,...,N \}$  where \\ $z_n=(z_{1n},z_{2n},...,z_{Jn})$, $z_{jn}=\hat{f}_{j}^{(-k)}(\boldsymbol{x}_n) = f^{(-k)}_{j} (\boldsymbol{x}_n,\hat{\boldsymbol{\theta}})$ , so $\hat{f}_{j}$ are predictions of level-0 learners on $\boldsymbol{x}_n $, for $j=1,2,3$.

\item Then, we will use a level-1 generalizer, or ensemble learner for predictions of a "new" data $\boldsymbol{x}$ on the testing dataset.
\begin{equation}
\resizebox{.85\hsize}{!}{$
\tilde{f}(\boldsymbol{x},\boldsymbol{\alpha})=  \alpha_{1}\hat{f}_{1}(\boldsymbol{x}) + \alpha_{2}\hat{f}_{2}(\boldsymbol{x}) + \alpha_{3}\hat{f}_{3}(\boldsymbol{x}) = \alpha_{1}f_{1}(\boldsymbol{x},\hat{\boldsymbol{\theta}}) + \alpha_{2}f_{2}(\boldsymbol{x},\hat{\boldsymbol{\theta}}) + \alpha_{3}f_{3}(\boldsymbol{x},\hat{\boldsymbol{\theta}})
$}
\end{equation}

\noindent Where the optimal $\hat{\boldsymbol{\theta}}$ weight vectors of models $f_j, j=1,2,3$ are estimated based on $L$ the whole historical data, and the weight vector $\boldsymbol{\alpha}$ is estimated using Equation (3) for a given $CV$.

\end{enumerate}

We call the learning, in Equation (3) and Equation (4), to determine the weight vector $\boldsymbol{\alpha}$, one-step learning, or $\alpha$ learning,  and denote the level-1 generalizer $\tilde{f}(\boldsymbol{x},\boldsymbol{\alpha})$ in Equation (4) as  $f_{E1}(\boldsymbol{x},\boldsymbol{\alpha})$ for the set of base models in $\{f_1, f_2, f_3$\}. So, the approximation $\hat{f}_{E1}$ for a new $\boldsymbol{x}$ based on optimal weight vector $\hat{\boldsymbol{\alpha}_1}$ for a given $CV$ is 

\[ 
\begin{split}
\hat{f}_{E1}(\boldsymbol{x})=f_{E1}(\boldsymbol{x}, \hat{\boldsymbol{\alpha}}_1) = \hat{\alpha}_{11}\hat{f}_1(\boldsymbol{x}) + \hat{\alpha}_{12}\hat{f}_2(\boldsymbol{x}) +  \hat{\alpha}_{13}\hat{f}_3(\boldsymbol{x}) \\ = 
 \hat{\alpha}_{11}f_{1}(\boldsymbol{x},\hat{\boldsymbol{\theta}}) + \hat{\alpha}_{12}f_{2}(\boldsymbol{x},\hat{\boldsymbol{\theta}}) + \hat{\alpha}_{13}f_{3}(\boldsymbol{x},\hat{\boldsymbol{\theta}})
 \end{split}
\]

As a result of the fourth learned base learner $\hat{f_4}$, $RBF$, we consider two additional standard approaches  $f_{E2}(\boldsymbol{x}, \boldsymbol{\alpha}_2)$, and $f_{E3}(\boldsymbol{x}, \boldsymbol{\alpha}_3)$, such that for a prediction of a new observation $x$ based on corresponding optimal weights $\hat{\boldsymbol{\alpha}}_1$ and $\hat{\boldsymbol{\alpha}}_2$ we have 

\[
\begin{split}
\hat{f}_{E2}(\boldsymbol{x})=f_{E2}(\boldsymbol{x}, \hat{\boldsymbol{\alpha}}_2) = \hat{\alpha}_{21}\hat{f}_1(\boldsymbol{x}) + \hat{\alpha}_{22}\hat{f}_3(\boldsymbol{x}) + \hat{\alpha}_{24}\hat{f}_4(\boldsymbol{x}) \\ 
= \hat{\alpha}_{21}f_{1}(\boldsymbol{x},\hat{\boldsymbol{\theta}}) + \hat{\alpha}_{22}f_{2}(\boldsymbol{x},\hat{\boldsymbol{\theta}}) + \hat{\alpha}_{24}f_{4}(\boldsymbol{x},\hat{\boldsymbol{\theta}})
\end{split}
\] 

\[ 
\begin{split}
\hat{f}_{E3}(\boldsymbol{x})=f_{E3}(\boldsymbol{x}, \hat{\boldsymbol{\alpha}}_3)) = \hat{\alpha}_{31}\hat{f}_1(\boldsymbol{x}) + \hat{\alpha}_{33}\hat{f}_3(\boldsymbol{x}) + \hat{\alpha}_{34}\hat{f}_4(\boldsymbol{x}) \\ = 
 \hat{\alpha}_{31}f_{1}(\boldsymbol{x},\hat{\boldsymbol{\theta}}) + \hat{\alpha}_{33}f_{3}(\boldsymbol{x},\hat{\boldsymbol{\theta}}) + \hat{\alpha}_{34}f_{4}(\boldsymbol{x},\hat{\boldsymbol{\theta}})
\end{split}
\] 

So, the ensembles $f_{E2}$, and $f_{E3}$ are linear combinations of base learners in the sets of $\{f_1, f_2, f_4$\}, and $\{f_1, f_3, f_4$\}, respectively. For the generation of the above ensemble learners we are keeping $f_1$ in each of possible combination of the the generated ensembles. We have found empirical evidence that for a few databases the $f_1$ model has a higher weight than any of the other models, giving the corresponding ensembles better prediction error than any of the individual models.

Therefore, all possible combinations of the remaining three level-0 learners taken two at a time makes a total of three level-1 learners: $f_{E1}$, $f_{E2}$ and $f_{E3}$ by keeping present $f_1$ in each ensemble.

\section{Level-2 Ensemble Learning}
\label{sec:4}
The motivation behind the level-2 ensemble learning we are proposing is to see whether combining, or mixing a set of level-1 learners using stacked generalization, we can improve the prediction error in comparison to the level-1 learning, particularly against the best level-1 learner among the standard stacking approaches. In the next two sections, we propose, as our first extension to standard stacking, two different level-2 learning approaches based on stacked generalization: the first approach, presented in Section 4.1, is based on two-step optimization, and the other, which is presented in Section 4.2, we call it all-at-once optimization. It is worth to mention that, the iterative nature of two-step learning will be advantageous over the all-at-once optimization to improve prediction error as we will discuss later at the end of this section.

\subsection{Two-step level-2 Learning}
\label{sec:4.1}
We will introduce our scheme of level-2 learning through iterative optimization based on optimal weights. Given the new predictions by the level-1 learners, $f_{E1}(\boldsymbol{x},\boldsymbol{\alpha}_1)$, $f_{E2}(\boldsymbol{x},\boldsymbol{\alpha}_2)$ and $f_{E3}(\boldsymbol{x},\boldsymbol{\alpha}_3)$, we can ask ourselves how to combine them? We suggest to apply the same $K-$fold $CV$ procedure on this new set of level-1 learners, instead of the ensemble composed of $f_1(\boldsymbol{x},\boldsymbol{\theta})$, $f_2(\boldsymbol{x},\boldsymbol{\theta})$, $f_3(\boldsymbol{x},\boldsymbol{\theta})$ and $f_4(\boldsymbol{x},\boldsymbol{\theta})$, to determine the corresponding set of weights $\beta_{1}$, $\beta_{2}$, and $\beta_{3}$, for $f_{E1}$, $f_{E2}$ and $f_{E3}$ respectively. The optimal weight vector $\hat{\boldsymbol{\beta}}$ is obtained through an optimization process, level-2 $generalizer$, resulting in corresponding set of optimal weights $\hat{\beta}_{1}$, $\hat{\beta}_{2}$ and $\hat{\beta}_{3}$. 

Once we get the optimal weight vector $\hat{\boldsymbol{\beta}}$, we can now combine the predictions of trained level-1 learners $f_{E1}(\boldsymbol{x},\hat{\boldsymbol{\alpha}}_1)$, $f_{E2}(\boldsymbol{x},\hat{\boldsymbol{\alpha}}_2)$ and $f_{E3}(\boldsymbol{x},\hat{\boldsymbol{\alpha}}_3)$ optimally as 
 $f_{E123}(\boldsymbol{x,\hat{\boldsymbol{\beta}}})=\hat{\beta}_{1}f_{E1}(\boldsymbol{x},\hat{\boldsymbol{\alpha}}_1)+\hat{\beta}_2f_{E2}(\boldsymbol{x},\hat{\boldsymbol{\alpha}}_2)+\hat{\beta}_{3}f_{E3}(\boldsymbol{x},\hat{\boldsymbol{\alpha}}_3)$ to get the approximation of level-2 learner.

It should be pointed out that if we expand, e.g.,  the equation for $f_{E123}$ according to 

\[f_{E1}=\alpha_{11}f_1 + \alpha_{12}f_2 + \alpha_{13}f_3 \]
\[f_{E2}=\alpha_{21}f_1 + \alpha_{22}f_2 + \alpha_{24}f_4 \]
\[f_{E3}=\alpha_{31}f_1 + \alpha_{33}f_3 + \alpha_{34}f_4 \]

\noindent
we get the following expression for $f_{E123}$ in terms of the models $f_1$, $f_2$, $f_3$ and $f_4$:

$f_{E123}(\boldsymbol{x},\boldsymbol{\gamma})=\gamma_{1}f_1(\boldsymbol{x},\boldsymbol{\theta)}+\gamma_{2}f_2(\boldsymbol{x},\boldsymbol{\theta})+\gamma_{3}f_3(\boldsymbol{x},\boldsymbol{\theta}) + \gamma_{4}f_4(\boldsymbol{x},\boldsymbol
{\theta})$ where 

$ \gamma_{1} = \beta_{1}\alpha_{11} + \beta_{2}\alpha_{21} + \beta_{3}\alpha_{31} $

$ \gamma_{2} = \beta_{1}\alpha_{12} + \beta_{2}\alpha_{22} $

$ \gamma_{3} = \beta_{1}\alpha_{13} + \beta_{3}\alpha_{33} $

$ \gamma_{4} = \beta_{2}\alpha_{24} + \beta_{3}\alpha_{34} $

Analysing any $\boldsymbol{\gamma}$'s weight, e.g. $\gamma_{2}$ corresponding to model $f_2$, we can observe that it is additively combining the $\boldsymbol{\alpha}$'s weights of model $f_2$, $\alpha_{12}$ and $\alpha_{22}$, which are present in ensembles $f_{E1}$ and $f_{E2}$ from first step of learning, through the extra weights $\beta_{1}$ and  $\beta_{2}$ that were gotten in the second step of learning for such ensembles.

Therefore, the set of weight vectors $\boldsymbol{\alpha}$ and $\boldsymbol{\beta}$ were obtained running two separated but iterative optimizations  which were called first step and second step of learning respectively.

We will denote this two-step optimization using the $\boldsymbol{\beta}$'s weights as follows:

\begin{itemize}

\item Given a dataset  $L=\{(y_n,\boldsymbol{x}_n ),n=1,...,N\}$ do $K$-fold $CV$. 

\begin{equation}
\min_{\boldsymbol{\beta}} \sum_{k} \sum_{(y_n,\boldsymbol{x}_n )\epsilon L_k} (y_n - \sum_{j} \beta_{j}f^{(-k)}_{Ej} (\boldsymbol{x}_n,\hat{\boldsymbol{\alpha}}_j))^2. 
\end{equation}

\[ \text{s.t. } \sum_j{\beta_j}=1.\]
\[\boldsymbol{\beta} \geq 0 \]

\end{itemize}

In the above equation, the $level$-2 data consists of $ \{(y_n,z_n),n=1,...,N \}$  where \\ $z_n=(z_{1n},z_{2n},...,z_{Jn})$, $z_{jn}=\hat{f}_{Ej}^{(-k)}(\boldsymbol{x}_n)=f^{(-k)}_{Ej} (\boldsymbol{x}_n,\hat{\boldsymbol{\alpha}}_j)$, and $\hat{f}_{Ej}$  on  $\boldsymbol{x}_n $ are predictions of level-1 learners for $j=1,2,3$.

We call this two-step learning, or $\beta$ learning, and denote the approximation, or level-2 ensemble learner based on weight vector $\boldsymbol{\beta}$ for predictions of a "new" data $x$ on the testing dataset as

\begin{equation}
\begin{split}
 f_{E123}(\boldsymbol{x},\boldsymbol{\beta}) =  \beta_{1}\hat{f}_{E1}(\boldsymbol{x}) + \beta_{2}\hat{f}_{E2}(\boldsymbol{x}) + \beta_{3}\hat{f}_{E3}(\boldsymbol{x}) \\ 
 = \beta_{1}f_{E1}(\boldsymbol{x},\hat{\boldsymbol{\alpha}}_1) + \beta_{2}f_{E2}(\boldsymbol{x},\hat{\boldsymbol{\alpha}}_2) + \beta_{3}f_{E3}(\boldsymbol{x}, \hat{\boldsymbol{\alpha}}_3)
\end{split}
\end{equation}

\noindent Where the optimal $\hat{\boldsymbol{\alpha}}_j$ weight vectors of models $f_{Ej}, j=1,2,3$ are estimated using an optimization similar to Equation (3) for a given $CV$, where we replace $ \hat{f}^{(-k)}_{j} \text{for} \hat{f}^{(-k)}_{Ej} $. On the other hand, the weight vector $\boldsymbol{\beta}$ is estimated using Equation (5) for a given $CV$.

\subsection{All-at-once level-2 Learning}
\label{sec:4.2}
We can as well run both optimizations at once where we optimize simultaneously both weight vectors $\boldsymbol{\alpha}$ and $\boldsymbol{\beta}$ which now will be denoted as $\boldsymbol{\alpha}^{'}$ and $\boldsymbol{\beta}^{'}$ respectively to emphasize that the optimized values are different from the iterative optimization. 

Therefore, given

$ \gamma^{'}_{1} = \beta^{'}_{1}\alpha^{'}_{11} + \beta^{'}_{2}\alpha^{'}_{21} + \beta^{'}_{3}\alpha^{'}_{31} $

$ \gamma^{'}_{2} = \beta^{'}_{1}\alpha^{'}_{12} + \beta^{'}_{2}\alpha^{'}_{22} $

$ \gamma^{'}_{3} = \beta^{'}_{1}\alpha^{'}_{13} + \beta^{'}_{3}\alpha^{'}_{33} $

$ \gamma^{'}_{4} = \beta^{'}_{2}\alpha^{'}_{24} + \beta^{'}_{3}\alpha^{'}_{34} $

\noindent
we can formulate such optimization via stacked generalization as follows:

\begin{multline*}
\min_{(\boldsymbol{\alpha}^{'},\boldsymbol{\beta}^{'})} \sum_{k} \sum_{(y_n,\boldsymbol{x_n})\epsilon L_k} (y_n - 
\{\gamma^{'}_{1}(\boldsymbol{\alpha}^{'},\boldsymbol{\beta}^{'})f_1(\boldsymbol{x_n},\hat{\boldsymbol{\theta}})+\gamma^{'}_{2}(\boldsymbol{\alpha}^{'},\boldsymbol{\beta}^{'})f_2(\boldsymbol{x_n},\hat{\boldsymbol{\theta}})  \\
+ \gamma^{'}_{3}(\boldsymbol{\alpha}^{'},\boldsymbol{\beta}^{'})f_3(\boldsymbol{x_n},\hat{\boldsymbol{\theta}}) + \gamma^{'}_{4}(\boldsymbol{\alpha}^{'},\boldsymbol{\beta}^{'})f_4(\boldsymbol{x_n},\hat{\boldsymbol{\theta}})\})^2
\end{multline*}

\[\text{s.t. } \alpha^{'}_{11}+\alpha^{'}_{12}+\alpha^{'}_{13}=1\]
\[ \alpha^{'}_{21}+\alpha^{'}_{22}+\alpha^{'}_{24}=1\]
\[ \alpha^{'}_{31}+\alpha^{'}_{33}+\alpha^{'}_{34}=1\]
\[ \beta^{'}_{1}+\beta^{'}_{2}+\beta^{'}_{3}=1\]
\[\alpha^{'} \geq 0 \]
\[\beta^{'} \geq 0 \]

\noindent
where 

$\alpha^{'}_{11}$, $\alpha^{'}_{12}$ and $\alpha^{'}_{13}$ are the set of weights for the set of level-0 learners $f_1, f_2, f_3$ in the learning ensemble $f^{'}_{E1}$.

$\alpha^{'}_{21}$, $\alpha^{'}_{22}$ and $\alpha^{'}_{24}$ are the set of weights for the set of level-0 learners $f_1, f_2, f_4$ in the learning ensemble $f^{'}_{E2}$.

$\alpha^{'}_{31}$, $\alpha^{'}_{33}$ and $\alpha^{'}_{34}$ are the set of weights of the level-0 learners $f_1, f_3, f_4$ in the learning ensemble $f^{'}_{E3}$.

and 
$\beta^{'}_{1}$, $\beta^{'}_{2}$ and $\beta^{'}_{3}$ are the set of weights for the level-1 learners $f_{E1}$, $f_{E2}$ and $f_{E3}$ in  learning ensemble $f^{'}_{E123}$.

This optimization, as we already said, optimize all parameters at once., and we call it $\gamma$ learning, and denote the approximation, or level-2 ensemble learner based on weight vector $\boldsymbol{\gamma}$ for predictions of a "new" data $x$ on the testing dataset as

$f^{'}_{E123}(\boldsymbol{x},\boldsymbol{\gamma})=\gamma^{'}_{1}f_1(\boldsymbol{x},\hat{\boldsymbol{\theta}})+\gamma^{'}_{2}f_2(\boldsymbol{x},\hat{\boldsymbol{\theta}})+\gamma^{'}_{3}f_3(\boldsymbol{x},\hat{\boldsymbol{\theta}}) + \gamma^{'}_{4}f_4(\boldsymbol{x},\hat{\boldsymbol{\theta}})$

\noindent Where the optimal $\hat{\boldsymbol{\theta}}$ weight vectors of models $f_j, j=1,2,3,4$ are estimated based on $L$ the whole historical data.

There is one more question that we should ask ourselves at this point. Is the two-step learning
better than all-at-once learning? This question could be answered in two ways. First by noting that according to the equivalence that we just showed for both methodologies, in terms of the corresponding gamma weights, we would expect to get very close results in terms of the prediction error for any database considered. Second, we found empirical evidence by running both methodologies across the database considered where we actually got very close results in terms of their corresponding error performances. However, the iterative nature of the two-step ensemble learning and its richness in structure in terms of of the pairwise correlation between each pair of level-1 ensemble predictions in the mixture, will be exploited and bring some insight about how to make two-step ensemble learning better than all-at-once optimization to improve error performance, as we show in detail in next section, particularly Section 5.2.

\section{Diversification and Selection for CV Partitions} 
\label{sec:5}
In stacked generalization the components of the ensemble, i.e. the level-0 learners, must be as diverse as possible, meaning that they provide predictions which are different from each other, otherwise if all of them provide the same predictions there will not be any improvements when we combine them.

In Section 5.1, we present in detail our second extension to standard stacking approach. As a first part of such extension, we propose to inject a diversity mechanism by perturbing the original $CV$ partition to generate new $CV$ partitions on which we can train new generated level-2 learners using stacked generalization, and two-step ensemble learning. Particularly, in Section 5.1.1, we introduce a combinatorial mechanism to generate all possible $CV$ partitions forming an exhaustive search space, as well as a systematic way to generate $CV$ partitions to shorten significantly the huge combinatorial search space given by the exhaustive search. In Section 5.1.2, we formulate two-step ensemble learning based on a new generated partition. 

In Section 5.2, we introduce, as a final part of the second extension, methodology we have developed for this work, that allows us to quantify the quality of a partition, and select one on which to make the final prediction. In Section 5.2.1, we exploit the structure of two-step level-2 ensemble learning to measure its diversity in terms of its base ensembles, or level-1 learners; introducing the concept of computed level-1 ensemble correlation in their prediction errors. This will provide us with  relevant diversity information about the quality of the different  $CV$ partitions. In Section 5.2.2, we introduce a decision matrix to support the selection of a partition.  In Section 5.2.3, we present a correlation-based criterion, and a partition ranking-based criterion to help us make a more informed decision by ranking such partitions. In Section 5.2.4, we introduce  heuristic rule-based Algorithm $S$, that uses the above two criteria,  to select the best $CV$ partition that contains a minimum correlation rank, with the aim to improve the prediction accuracy of the level-2 ensemble corresponding to the original $CV$. 

\subsection{Diversification through Generation of CV partitions}
\label{sec:5.1}
\subsubsection{Exhaustive and Systematic Generation of Partitions}
\label{sec:5.1.1}
In our initial $K$-fold $CV$ was obtained by removing one of the $K$ subsets, which results in a set of partitions, that will be represented by  $CV^0$. In order to generate different subsets of learning datasets  we now take out two subsets instead of only one fold out of $K$. The total number of possibly different partitions sets that can be generated  this way out of a $K$-fold $CV$ partition is prohibitively large, with a total of $M={(K)}^K-1$ new partitions in addition to the original $CV^0$ partition. So, we will generate only $M=K$ new different $CV$ partitions in an orderly way. It will provide us with a systematic mechanism to explore a reduced set of alternatives, and to test the algorithms proposed in this paper in a computationally efficient way. For some $m$ in $\{1,2,...,M\}$, the partition set for a new systematic generated  partition will be denoted as $CV^{m}$. Please see Section 6.1 for details of the implementation of the new partitions. 

\subsubsection{Two-Step ensemble learning based on  $CV^{m}$ partition}
\label{sec:5.1.2}

Given the partition set of corresponding generated partition $CV^{m}$, we have the following formulation of two-step ensemble learning:

\begin{itemize}

\item Given a dataset  $L=\{(y_n,\boldsymbol{x}_n ),n=1,...,N\}$ do $K$-fold $CV^{m}$. 

\begin{equation}
\min_{\boldsymbol{\beta}} \sum_{k} \sum_{(y_n,\boldsymbol{x}_n )\epsilon L_k} (y_n - \sum_{j} \beta_{j}f^{(-k)(-m))}_{Ej} (\boldsymbol{x}_n,\hat{\boldsymbol{\alpha}}_j)^2. 
\end{equation}

\[\text{s.t.} \sum_j{\beta_{j}}=1.\]
\[\boldsymbol{\beta} \geq 0 \]

\end{itemize}

We denote the level-2 ensemble learner based on weight vector $\beta_{m}$, for predictions of a "new" data $x$ on hold-out dataset, given $CV^{m}$  as 

\begin{equation}
\begin{split}
f_{E123^{m}}(\boldsymbol{x},\boldsymbol{\beta}) = \beta_{1}\hat{f}_{E1^{m}}(\boldsymbol{x}) + \beta_{2}\hat{f}_{E2^{m}}(\boldsymbol{x}) + \beta_{3}\hat{f}_{E3^{m}}(\boldsymbol{x}) \\
=  \beta_{1}f_{E1^{m}}(\boldsymbol{x},\hat{\boldsymbol{\alpha}}_1) + \beta_{2}f_{E2^{m}}(\boldsymbol{x},\hat{\boldsymbol{\alpha}}_2)+ \beta_{3}f_{E3^{m}}(\boldsymbol{x},\hat{\boldsymbol{\alpha}}_3)
\end{split}
\end{equation}

Equation (7) and Equation (8), are equivalent to Equation (5) and Equation (6) respectively. In Equation (8) we denote with an index $m$ the dependency on a given $CV^{m}$ partition of level-1 learners $f_{Ej^{m}}$, with $j=1,2,3$, and  the level-2 learner $f_{E123^{m}}$ as well. The reason of this is that, the $\hat{\boldsymbol{\alpha}}$ weight vectors of level-1 learners present in Equation (7)
is estimated using an optimization similar to Equation (3), but in the context of a given $CV^m$ partition, and the weight vector $\boldsymbol{\beta}$ in Equation (8) is estimated as well for a given $CV^{m}$ by solving Equation (7). In Section 6.1, we explain in a little bit more detail the two-step ensemble learning.

\subsection{Ensemble Correlation-based Selection of Appropriate CV Partition}
\label{sec:5.2}
In Section 5.2.1, we will introduce some concepts to help us define a level-1 ensemble correlation-based criterion, which will provide relevant information about the quality of a cross-validation partition. In Section 5.2.2, we introduce a decision matrix which will facilitate the way we make the selection of an appropriate partition. In Section 5.2.3, we present a correlation-based criterion, and a partition ranking-based criterion that can be used to rank partitions. In Section 5.2.4, both criteria are employed by Algorithm $S$, which is a max-min rule-based systematic learning algorithm, to select best $CV$ partition, that contains a minimum correlation rank.  

\subsubsection{Basic Concepts}
\label{sec:5.2.1}
Next, we will introduce some notation that we will use in the description of the selection criteria in the next two sections.

In the framework of two-step level-2 learning represented by Equation (8), let us look at the level-2 data which are input for the optimization process in Equation (7), i.e., the actual values $y_n$ along with predictions $\hat{f}^{(-k)(-m)}_{Ej}(\boldsymbol{x}_n)=f^{(-k)(-m)}_{Ej}(\boldsymbol{x}_n,\hat{\boldsymbol{\alpha}}_j)$, for $J=1,2,..., J$. The actual values $y_n$ along with the above predictions collected from each of the $K$ validation folds of $CV^m$ are used to define the error matrix as

\[\boldsymbol{E}_{CV^m}=(\boldsymbol{e}^m_1, \boldsymbol{e}^m_2,...,\boldsymbol{e}^m_J)=
\begin{pmatrix}
	e^m_{11} & e^m_{12} & e^m_{1J} \\
	e^m_{21} & e^m_{22} & e^m_{2J} \\
	                         \\
	e^m_{N1} & e^m_{N2} & e^m_{NJ} 
	
\end{pmatrix}
\]	

\noindent where each element in $\boldsymbol{E}_{CV^m}$ is defined as $e^m_{nj} = y_n - \hat{f}^{(-k)(-m)}_{Ej}(\boldsymbol{x}_n);$ $j=1,...,J, n=1,..., N,$ and $(\boldsymbol{x}_n,y_n) \in L_k$ with $k=1,..., K.$

From the error matrix $\boldsymbol{E}_{CV^m}$, the sample variance-covariance matrix between columns of size $J \times J$: $  \boldsymbol{V}_{CV^m}=(V^m_{jl})$ is calculated as well as its sample correlation matrix between columns $ \boldsymbol{R}_{CV^m}=(r_{jl}^m)$.

In our prediction problem matrix $\boldsymbol{R}_{CV^m}$ is symmetric, and the diagonal elements are equal to one, as it is the correlation between the predictions of level-1 learners  $f_{Ej}$ and $f_{El}$. We are interested in the $(J-1)J/2$ elements  above the diagonal of such matrix, and we put them in a row-wise fashion into the following row vector $\textbf{r}_{_{CV^m}}$: 
\begin{equation} 
          \boldsymbol{g.}_m=\textbf{r}_{_{CV^m}}=(r^{m}_{12},r^{m}_{13},...,r^{m}_{1J},r^{m}_{23},...,r^{m}_{2J},r^{m}_{34},...,r^{m}_{(J-1)J})
                    =(r^{m}_{jl})_{1 \leq j<l \leq J} 
\end{equation} 

In the last equation we are representing correlation vector $\textbf{r}_{\textbf{cv}^m}$ by an alternative vector $\boldsymbol{g.}_m$  containing $ Q=(J-1)J/2$ criteria, or elements, where $\boldsymbol{g.}_m = (g_1(\boldsymbol{g.}_m), g_2(\boldsymbol{g.}_m),...,\allowbreak g_{Q}(\boldsymbol{g.}_m))=(g_{1m},g_{2m},..., g_{Qm}) = (g_{qm})_{q=1,2,..., Q}$. Each criterion $g_q$ applied on an alternative $\boldsymbol{g.}_m$ and represented for element $g_{qm}$ in vector $\boldsymbol{g.}_m$ with $q=1,...,Q$, measures the performance of alternative $\boldsymbol{g.}_m$ on criterion $g_q$, and corresponds to an error correlation function between a pair of ensemble, or level-1 learner predictors, $f_{Ej}$ and $f_{El}$ with $1 \leq j < l \leq J$ in vector   $\textbf{r}_{_{CV^m}}$. So, when we talk about  partition $CV^{m}$ we will be referring to alternative $\boldsymbol{g.}_m$, and vice-versa. 

It should be emphasized that it is here where we exploit the structure of two-step level-2 ensemble algorithm as opposed to all-at-once level-2 learning, which lacks such structure by construction. At the end of the first step of two-step level-2 learning, or level-1 learning, and for a given $K$-fold $CV^m$ partition, we can compute the error correlations $r_{jl}$, before the optimization in Equation (7), between the prediction errors of level-1 learner approximations $f^{(-k)(-m)}_{Ej}(\boldsymbol{x}_n,\hat{\boldsymbol{\alpha}}_j)$ and $f^{(-k)(-m)}_{El}(\boldsymbol{x}_n,\hat{\boldsymbol{\alpha}}_j)$ where $\boldsymbol{x}_n  \in L_k$, and  $L_k$ with $k=1,2,..., K$ are validation folds. 
This correlation information provides relevant information about the quality of the $K$-fold $CV^m$ partition as we can see in vector  $\textbf{r}_{_{CV^m}}$ from Equation (9), and it is the basis for Algorithm $S$ that we present in the next sections.

\subsubsection{Decision Matrix for Selection of a $CV$ partition}
\label{sec:5.2.2}
In this section we build a decision matrix which will facilitate the way we make the selection of an appropriate level-2 learner. The matrix will have as rows alternatives corresponding to the different generated $CV$ partitions.

In the general case, if we have $M+1$ different alternatives, $\boldsymbol{g.}_m$  with $m=0,1,...,M$ , which are associated to partitions $CV^{m}$ with $m=0,..., M$, and $Q$ criteria $g_{q}$, $q=1,2,...,Q$ we can build the following decision matrix $\boldsymbol{G}$: 

$$\bordermatrix{\text{ } &
g_{1}  & g_{2}  & \ldots  & g_{Q}\cr 
\boldsymbol{g.}_0 & g_{10} & g_{20} & \ldots & g_{Q0}\cr 
\boldsymbol{g.}_1 & g_{11} & g_{21} & \ldots & g_{Q1}\cr 
      \ldots  & \ldots & \ldots & \ldots & \ldots \cr      
\boldsymbol{g.}_m & g_{1M} & g_{2M} & \ldots & g_{QM} }
$$

\noindent Where $g_{qm}$ is the performance of alternative $\boldsymbol{g.}_m$ on criteria $g_q$, for $q=1,...,Q$.

We can determine the minimum correlation set $G^{min}$ as the line vector of minimums column-wise. 

We will refer alternatively to the decision matrix  $\boldsymbol{G}$ as the set  of alternatives, i.e., we will also use the notation:
 $\boldsymbol{G} = \{\boldsymbol{g.}_0, \boldsymbol{g.}_1, \boldsymbol{g.}_2, ...,\boldsymbol{g.}_m \}$.

From now on, when using a set $\boldsymbol{G}$ of alternatives we will be making reference to the systematic generation of alternatives, $\boldsymbol{g.}_m$ with $m = 0,1,,..., M$, where $M$, the number of new alternatives, is equal to the number of $K$ of folds by design of systematic learning., i.e., $M=K$. It should be emphasized that  each alternative $\boldsymbol{g.}_m$ in $\boldsymbol{G}$ is related to a corresponding $K$-fold $CV^{m}$ partition, as vector $\boldsymbol{g.}_m$ is equal to vector $\textbf{r}_{\textbf{cv}^m}$ according to Equation (9).

On the other hand, in Equation (8), level-2 learning, we make reference to a level-2 generalizer, or learner $f_{E123^{m}}$ for a $CV^{m}$, for some $m$ in $\{ 0,1,2,..., M\}$. We use $f_{E123^{m}}$ for prediction of a new instance $\boldsymbol{x}$. In section 5.2.3, we present in detail two rule-based criteria used for selection of a $CV^m$ partition. These two rules are employed within Algorithm $S$ in Section 5.2.4 to choose one alternative $\boldsymbol{g.}_m$ for some $m$ in $\{0,1,2,..., M\}$. Once we have such alternative we can use its corresponding level-2 learner $f_{E123^{m}}$, to make predictions of new instances  $\boldsymbol{x}$. One of such rules is a correlation based criterion  between predictions of level-1 learners, while the other rule is based on a rank-based criterion of such correlations.  

\noindent
\subsubsection{Partition Selection using Correlation-based Criterion and Ranking-based Criterion}
\label{sec:5.2.3}
\noindent

In \citet{keung06} the authors use an  uncorrelation maximization algorithm, which is based on a correlation based criterion, to remove ensemble learners based on a threshold $\theta=0.8$ with the objective of determining the best subset of ensemble learners. Their algorithm is based on the principle of model diversity, where the correlation between the predictors should be as small as possible \citep{keung06}. In this section we will introduce a computed level-1 ensemble correlation criterion, and a rank based criterion, not to prune an ensemble, but to handle additional, and diverse $CV$ partitions, rank them and select one. The selected partition has to contain a minimum correlation rank as we will explain in Algorithm $S$. One of its advantages is that they do not depend on extra parameter values like $\theta$.

The basis of the correlation criterion to select a particular $CV^m$ partition, is to compute for a two-step level-2 ensemble, $f_{E123m}$, or corresponding partition $CV^{m}$, the error correlation between its level-1 learner predictions $ \hat{f}^{(-k)(-m)}_{Ej}$ at the end of level-1 learning step, for $j= 1,2,3$, represented by alternative $\boldsymbol{g.}_m$ in $\boldsymbol{G}$. The correlations, measured at this extra level of smoothing in the predictions, gives a way to control the diversity of its base ensembles, or level-1 learner components, and therefore be able to quantify the quality of a partition using a correlation based criterion, and a correlation rank based criterion.

In next section we introduce Algorithm $S$, where we select the best alternative by grouping alternatives into different segments. For the first segment of alternatives we use a Median based selection rule, $Med$ rule, and in case we are not able to do a selection due to potential outliers in such alternative, then we go to next segment where we use $FindBestAR$ rule for such selection instead. If we determine that there are potential outliers in such alternatives, then we go to next segment and apply $FindBestAR$ rule again, until we eventually select one with no potential outliers, or if we ran out of alternatives then we keep the original alternative $\boldsymbol{g.}_0$ corresponding to original partition $CV^{0}$.  

We will describe some of the inputs required for Algorithm $S$. First, $G^{min}$  is the set of minimum correlations collected across all $q$-th criteria, $g_q$ in $\boldsymbol{G}$; $G^{min}$ is then used for $Med$ rule to make a selection, as we will describe in next section.

Algorithm $S$ also requires, a set $\boldsymbol{T}$ of rank vectors of alternatives, and a set $\boldsymbol{ar}$ of scores of alternatives, which we define below, and are used by the $FindBestAR$ rule in Algorithm $S$ to make a selection. These inputs are generated before hand by the Average Ranking Method, which we describe next briefly.

The Average Ranking Method $(AR)$ \citep{bentley98} selects one criterion, $g_q$ of $\boldsymbol{G}$, which contains a set of alternative's performances: $g_q(\boldsymbol{g.}_{m}) = g_{qm}$, for $m=0,1,..., M$. Then, it computes a set of rankings on them to get $rank(g_q(\boldsymbol{g.}_0)), rank(g_q(\boldsymbol{g.}_1)),\allowbreak ...,  rank(g_q(\boldsymbol{g.}_m))$, where $rank(g_q(\boldsymbol{g.}_m)) $, returns the rank of $\boldsymbol{g.}_{m}$ in a set of alternatives $\boldsymbol{g.}_{0}$, $\boldsymbol{g.}_{1}$, $ ..., \boldsymbol{g.}_{M}$ according to the $g_q$ criterion. Rank value $1$ means the best criterion value, and rank value $M+1$ means the worst criterion value. It repeats this procedure for criterion $g_q$, with $q=1,2,..., Q$.

The ranking for an alternative $\boldsymbol{g.}_{m}$ is then given by  the rank vector
\[ \boldsymbol{rank}(\boldsymbol{g.}_{m}) = (rank(g_1(\boldsymbol{g.}_m)), rank(g_2(\boldsymbol{g.}_m)), ..., rank(g_Q(\boldsymbol{g.}_m))   )\]

Once the rank vector $\boldsymbol{rank}(\boldsymbol{g.}_{m})$ is computed for each alternative $\boldsymbol{g.}_{m}$ in $\boldsymbol{G}$, its score is calculated by adding the ranks of $\boldsymbol{g.}_{m}$ for each criterion. 

\[  ar({\boldsymbol{g.}_m}) = \sum_{q=1}^{Q}  rank(g_q(\boldsymbol{g.}_m))  \]

Let $\boldsymbol{T}$ be the set of rank vectors:

\[
 \boldsymbol{T} = \{	\boldsymbol{rank}(\boldsymbol{g.}_{0}), \boldsymbol{rank}(\boldsymbol{g.}_{1}), ...,\boldsymbol{rank}(\boldsymbol{g.}_{M}) \}
\]

And $\boldsymbol{ar}$ the set of alternative scores \[ \boldsymbol{ar} = \{ ar({\boldsymbol{g.}_0}),  ar({\boldsymbol{g.}_1}), ...,  ar({\boldsymbol{g.}_{M}})  \} \]

Next, we give the specification of Algorithm $S$ in terms of its inputs, and outputs:

\[ [\boldsymbol{g.}^*] \leftarrow S(\textit{M+1},Q,\boldsymbol{G},G^{min},\boldsymbol{T},\boldsymbol{ar} )\]

So, Algorithm $S$ returns best alternative. Then, we can predict with corresponding best level-2 learner $f_{E123^*}$  for a new instance $\boldsymbol{x}$: $f_{E123^*}(\boldsymbol{x},\boldsymbol{\beta)}$.

\subsubsection{Rule-based Systematic Ensemble Learning Algorithm for Ensemble Selection}
\label{sec:5.2.4}
In a high level description, the main objective of Algorithm $S$ is to consider alternatives in $\boldsymbol{G}$ containing a minimum rank in $\boldsymbol{T}$, and select one of them by using one of two selection rules. In order to do that we proceed at different segments, we first find all rank vectors of alternatives in  $\boldsymbol{T}$ with $rank=1$, or $currentsegment=1$, in at least one of its coordinate elements. Using a selection rule at $segment=1$, $Med$ selection rule, an alternative is selected  as long as such alternative is not disregarded when applying two rules used to detect potential outliers in it. Otherwise, all alternatives with $rank=1$ are eliminated from $\boldsymbol{T}$, and we then proceed to next segment, $currentsegment=currentsegment+1$, by finding all alternatives with a minimum rank, $rank=currentsegment$, and put them in updated set $\boldsymbol{T}$. A second selection rule, $BestFindAR$ is used at this segment to select an alternative with minimum score as long as such alternative is not disregarded if outliers are not detected in it. Otherwise, the disregarded alternative is eliminated from set $\boldsymbol{T}$, and repeat the same steps at same $currentsegment$; applying $BestFindAR$ rule on remaining alternatives in $\boldsymbol{T}$. If we run out of elements in $\boldsymbol{T}$, then we go to next segment by increasing $currentsegment$ by 1, finding all alternatives with next minimum rank, and applying $BestFindAR$ rule until either an alternative is selected with no potential outliers at $currentsegment$, or either  we reach the maximum number of segments $M$, or there is not any alternative left in $\boldsymbol{T}$. In any of these two last cases is true then we keep original partition for final prediction. 

It is worth to mention that at $currentsegment=1$ we applied $Med$ Rule as opposed to had applied $FindBestAR$, because the former rule is more robust than the latter rule based on empirical evidence through the databases considered in this work. In addition to that, we believe that at $segment=1$ we have an extreme case where the rank vectors have at least one of its element with minimum rank, $rank=1$ across all rank vector available, and the $Med$ Rule, takes into account extra information, where it selects from $\boldsymbol{G}$ a compromised alternative containing the median of the minimum correlation values, which corresponds to one of the rank vectors with $rank=1$. On the other hand, at the upper segments, $rank>1$, we do not have that extreme case any more, and there is not need to use the more specialized $Med$ rule that takes into account correlation information explicitly. 
Next, we describe in more detail Algorithm $S$, focusing more in the selection rules, and two rules designed to help us detect an alternative with potential outliers. First rule, checks if an alternative contains two rank values with a minimum value at $currentsegment$, the second rule detects if such alternative has a maximum rank value equal to $M+1$. 
\\
\indent In the first part, labelled $Median$ $Rule$, lines 1-14, where $currentsegment=1$, the objective is to apply $Med$ Rule on $\boldsymbol{G}$ in order to get an alternative $\boldsymbol{g.}_{cen}$, or center alternative,  containing the median value in set $G^{min}$, with no potential outliers in it. We mark criterion $g_{q*}$, which contains such minimum median value, where $rank(g_{q*}(\boldsymbol{g.}_{cen}))=1$. In line 7 and line 8, we apply $Min$ and $Max$ rules on  $\boldsymbol{rank}(\boldsymbol{g.}_{cen})$ to help us find minimum and maximum rank values in it, respectively. Those values will help us to determine whether such alternative has any potential outliers. On one hand, $Min$ rule finds a criterion $g_{i*}$ in  $\boldsymbol{rank}(\boldsymbol{g.}_{cen}) \setminus rank(g_{q*}(\boldsymbol{g.}_{cen}))$ with minimum rank: $rank(g_{i*}(\boldsymbol{g.}_{cen}))$. On the other hand, $Max$ rule finds a criterion $g_{j*}$ in  $\boldsymbol{rank}(\boldsymbol{g.}_{cen})$ with maximum rank: $rank(g_{j*}(\boldsymbol{g.}_{cen}))$.  In line 10, we check if the center rank vector has two minimum rank values, i.e., $rank(g_{i*}(\boldsymbol{g.}_{cen})) = rank(g_{q*}(\boldsymbol{g.}_{cen})) = 1 = curentsegment \text{ with } g_{i*} \neq g_{q*}$, or if its maximum rank value $rank(g_{j*}(\boldsymbol{g.}_{cen}))$ is equal to $M+1$. If any or both of these conditions are true then we go to second part of the algorithm, as we have found an alternative with potential outliers. Otherwise, if both conditions are false then we select $\boldsymbol{g.}_{cen}$ as $\boldsymbol{g.}^{*}$, and go to line 36, and return alternative $\boldsymbol{g.}^{*}$, whose corresponding two-step level-2 ensemble, $f_{E123^*}$, will be  used to make predictions of new instances $\boldsymbol{x}$. 
\\
\indent In the second part, line 15 to line 35, labelled $Average$ $Ranking$ $Rule$, we first go to next segment by increasing $currentsegment$ by 1, and collect all rank vectors  from  $\boldsymbol{T}$ with $rank=currentsegment$ into $\boldsymbol{T1}$. The objective is to find a best alternative $\boldsymbol{g.}_{best}$ with minimum $ar$ score, or corresponding best rank vector $\boldsymbol{rank}(\boldsymbol{g.}_{best})$ from $\boldsymbol{T1}$, that has minimum rank, i.e., $rank(g_{q*}(\boldsymbol{g.}_{best}))=currentsegment$, with no potential outliers in it. So, we apply again the same $Min$ and $Max$ rules, but now on $\boldsymbol{rank}(\boldsymbol{g.}_{best})$. In line 26, we check whether the best rank vector has two minimum rank values at $currentsegment$, or if its maximum rank value is equal to $M+1$. If any or both of these conditions are true then we go on in second part of the algorithm. Otherwise, if both conditions are false then we select $\boldsymbol{g.}_{cen}$ as $\boldsymbol{g.}^{*}$, and go to line 36, and return alternative $\boldsymbol{g.}^{*}$, whose corresponding two-step level-2 ensemble, $f_{E123^*}$, will be used to make predictions of new instances $\boldsymbol{x}$.  

\begin{algorithm}
%Remove Algorithm #
\renewcommand{\thealgorithm}{}
\caption{S: Max-min rule-based Systematic Ensemble Learning Algorithm for Selection}
{\fontsize{10}{10}\selectfont
\begin{algorithmic}[1]

\REQUIRE
\begin{math}
\textit{M+1} \text{ } \text{[number of alternatives],} \textit{Q} \text{ } \text{[number of criteria] }     
\end{math}	

\REQUIRE
\begin{math}
\boldsymbol{G} \text{ } \text{is set of alternatives} \text{ }
\{	\boldsymbol{g.}_{0}, \boldsymbol{g.}_{1}, \boldsymbol{g.}_{2}, ...,\boldsymbol{g.}_{M} \} 
\end{math}

\REQUIRE{$G^{min}$ [Set of minimum correlation values of each criterion $g_q$]   
 }\\
\REQUIRE
\begin{math}
\boldsymbol{T} \text{ } \text{is set of rank vectors } \text{ }
 \{	\boldsymbol{rank}(\boldsymbol{g.}_{0}), \boldsymbol{rank}(\boldsymbol{g.}_{1}), \boldsymbol{rank}(\boldsymbol{g.}_{2},; ...,\boldsymbol{rank}(\boldsymbol{g.}_{M}) \}   
\end{math}	

\begin{math}
\text{ where }
\resizebox{.88\hsize}{!}{$
 \boldsymbol{rank}(\boldsymbol{g.}_{m}) \text{ } \text{is vector of alternative ranks } \text{ }
  (rank(g_1(\boldsymbol{g.}_m)), rank(g_2(\boldsymbol{g.}_m)), ..., rank(g_Q(\boldsymbol{g.}_m))     )
  $}
\end{math}	

\REQUIRE $\boldsymbol{ar} \text{ } \text{is a set of} \text{ } \text{alternative scores} \text{ }  \{ ar({\boldsymbol{g.}_0}),  ar({\boldsymbol{g.}_1}), ...,  ar({\boldsymbol{g.}_{M}})  \} $

\COMMENT {Part 1 - Median Rule } % single line comment
\STATE{ $ \boldsymbol{T1}, \boldsymbol{G1} \leftarrow \emptyset, \boldsymbol{g.}^{*} \leftarrow \boldsymbol{g.}_{0}, currentsegment \leftarrow  1 $ }
\FORALL {$\boldsymbol{rank}(\boldsymbol{g.}_m) \in \boldsymbol{T}$ with $rank(g_i(\boldsymbol{g.}_m)) = currentsegment=1$, for some $i$ }
\STATE{  $\boldsymbol{T1} \leftarrow \boldsymbol{T1} \cup \{\boldsymbol{rank}(\boldsymbol{g.}_m)\}, \boldsymbol{G1} \leftarrow \boldsymbol{G1} \cup \{\boldsymbol{g.}_m$\} } 
\ENDFOR
\STATE{$\boldsymbol{g.}_{cen} \leftarrow Med(\boldsymbol{G},Q,G_{min}) $} [Find a minimum correlation alternative based on Median Rule]
\STATE {Mark $\boldsymbol{g.}_{cen}$, rank vector $\boldsymbol{rank}(\boldsymbol{g.}_{cen})$, and criterion $g_{q*}$ where $rank(g_{q*}(\boldsymbol{g.}_{cen}))=1$ }

\STATE {\textit{Min Rule:} $g_{i*} \leftarrow \argmin\limits_{ g_{i} \in \{g_{1},...,g_{Q} \} \setminus g_{q*} }  \{rank(g_{i}(\boldsymbol{g.}_{cen} )) \} $} 
[Find criterion $g_{i*}$ with minimum rank, $rank(g_{i*}(\boldsymbol{g.}_{cen} ))$, in center rank vector excluding rank of $g_{q*}(\boldsymbol{g.}_{cen})$ from it ]

\STATE {\textit{Max Rule:} $g_{j*} \leftarrow \argmax\limits_{ g_{j} \in \{g_{1},...,g_{Q} \} } \{rank(g_{j}(\boldsymbol{g.}_{cen} )) \} $ } [Find $g_{j*}$ with maximum rank  ]

\IF{$(rank(g_{i*}(\boldsymbol{g.}_{cen}))= 1 \text{ where } g_{i*} \neq g_{q*}) \boldsymbol{\lor} rank(g_{j*}(\boldsymbol{g.}_{cen}))= \text{M+1} $} 
\STATE {Unmark $\boldsymbol{g.}_{cen}$,  $\boldsymbol{rank}(\boldsymbol{g.}_{cen})$ and $g_{q*}$. }[ Unmark alternative that has potential outliers]
\ELSE{ }
\STATE {$\boldsymbol{g.}^* \leftarrow  \boldsymbol{g.}_{cen}$ \AND Go to line 36}
\ENDIF

\STATE {$\boldsymbol{G} \leftarrow \boldsymbol{G} \setminus \boldsymbol{G1}$} \AND {$\boldsymbol{T} \leftarrow \boldsymbol{T} \setminus \boldsymbol{T1}$} [Update $\boldsymbol{G}$ and $\boldsymbol{T}$ eliminating alternatives with $rank = 1$]
\\\hrulefill
\\
\COMMENT {Part 2 - Average Ranking Rule} % single line comment
\STATE $currentsegment \leftarrow currentsegment + 1$  
\FOR  {$currentsegment = 2$ to $currentsegment=M$ \AND $\boldsymbol{T} \neq \emptyset$ }
\STATE $\boldsymbol{T1}, \boldsymbol{ar1} \leftarrow \emptyset $ 

\FORALL {$\boldsymbol{rank}(\boldsymbol{g.}_m) \in \boldsymbol{T}$ with $rank(g_i(\boldsymbol{g.}_m)) = currentsegment$, for some $i$ }
\STATE{  $\boldsymbol{T1}=\boldsymbol{T1} \cup \{\boldsymbol{rank}(\boldsymbol{g.}_m)\}, \boldsymbol{ar1} =\boldsymbol{ar1} \cup \{ar(\boldsymbol{g.}_m)$\} } 
\ENDFOR

\WHILE {$\boldsymbol{T1} \neq \emptyset$}
\STATE  $\boldsymbol{rank}(\boldsymbol{g.}_{best}) \leftarrow FindBestAR( \boldsymbol{T1},\boldsymbol{ar1},currentsegment)$ [Find best rank vector].
\STATE {Mark $\boldsymbol{g.}_{best}$, $\boldsymbol{rank}(\boldsymbol{g.}_{best})$, and criterion $g_{q*}$ where $rank(g_{q*}(\boldsymbol{g.}_{best}))=currentsegment$ }

\STATE {\textit{Min Rule:} $g_{i*} \leftarrow \argmin\limits_{ g_{i} \in \{g_{1},...,g_{Q} \} \setminus g_{q*} }  \{rank(g_{i}(\boldsymbol{g.}_{best} )) \} $} 
[Find criterion $g_{i*}$ with minimum rank, $rank(g_{i*}(\boldsymbol{g.}_{best} ))$, in best rank vector excluding rank of $g_{q*}(\boldsymbol{g.}_{best})$ ]

\STATE {\textit{Max Rule:} $g_{j*} \leftarrow \argmax\limits_{ g_{j} \in \{g_{1},...,g_{Q} \} } \{rank(g_{j}(\boldsymbol{g.}_{best} )) \} $} [Find $g_{j*}$ with maximum rank]

\IF{$(rank(g_{i*}(\boldsymbol{g.}_{best}))=\textit{currentsegment} \text{ where } g_{i*} \neq g_{q*}) \boldsymbol{\lor} rank(g_{j*}(\boldsymbol{g.}_{best}))= \text{M+1} $}
\STATE {Unmark $\boldsymbol{g.}_{best}$ ,  $\boldsymbol{rank}(\boldsymbol{g.}_{best})$ and $g_{q*}$.}[Unmark alternative that has potential outliers]
\STATE { $ \boldsymbol{T1} = \boldsymbol{T1} \setminus \{\boldsymbol{rank}(\boldsymbol{g.}_{best})\}$} \AND  
{$\boldsymbol{ar1} = \boldsymbol{ar1} \setminus \{ar(\boldsymbol{g.}_{best})\}$}
\STATE {$\boldsymbol{G} = \boldsymbol{G} \setminus \{\boldsymbol{g.}_{best}\}$} \AND {$\boldsymbol{T} = \boldsymbol{T} \setminus \{\boldsymbol{rank}(\boldsymbol{g.}_{best})\}$} 
\ELSE{}
\STATE {$\boldsymbol{g.}^* \leftarrow  \boldsymbol{g.}_{best}$ \AND Go to line 36}
\ENDIF

\ENDWHILE
\STATE  $currentsegment \leftarrow currentsegment + 1$
\ENDFOR 

%%%%%%%%%%%%%%%%%%%%%%%%%%%%%%%%%%%%%%%%%%%%%%%%%%%%%%%%%%%%%

\RETURN {$ \boldsymbol{g.}^*$} [Return best alternative, whose level-2  learner $f_{E123^*}$ will be used for prediction]\\ 

\end{algorithmic}
}
\end{algorithm}

\pagebreak

\section{Experimental Set-up}
\label{sec:6}
\subsection{Implementation Details}
\label{sec:6.1}
In our initial $K$-fold $CV$ we partitioned the original historical dataset $L$ in the following disjoint sub-samples $L_k$ with $k=1,...,K$. Then, we built the learning or training subsets $L^{(-k)}$ by  where $L^{(-k)} = L - L_{k}$, and we denoted $L_{k}$ as the validation folds.  

The above results in a set of partitions $(L^{(-1)}, L_{1}), (L^{(-2)}, L_{2}), ..., (L^{(-K)}, L_{K}) $, where $L^{(-k)}$ are the training subsets, and $L_k$ the validation folds, with $k=1,2,..., K$ . Making an abuse of notation, such set will be represented as:

\[CV^0 = \{(L^{(-1)}, L_{1}), (L^{(-2)}, L_{2}), ..., (L^{(-K)}, L_{K}) \} \]

In order to generate different subsets of learning datasets  we now take out two folds instead of only one fold. 
The total number of possibly different partitions sets that can be generated  this way out of a $K$-fold $CV$ partition is prohibitively large, with a total of $M={(K)}^K-1$ new partitions in addition to the original $CV^0$ partition. The method proposed next generates only $M=K$ new different $CV$ partitions in an orderly way. It will provide us with a systematic mechanism to explore a reduced set of alternatives, and to test the algorithms proposed in this paper in a computationally efficient way.  It should be noted that for the systematic procedure the number of partitions $M$ coincides with $K$, the number of folds, i.e., $M=K$.  

For some $m$ in $\{1,2,...,M\}$, and e.g., $K=5$, the partition set for a new systematic generated  $CV^{m}$ partition is 

\[
\begin{split}
CV^m=\{(L^{(-1)(-m)}, L_{1}), (L^{(-2)(-m)}, L_{2}), (L^{(-3)(-m)},L_{3}), \\ (L^{(-4)(-m)}, L_{4}),(L^{(-5)(-m)}, L_{5}) \}
\end{split}
\]

\noindent where $L^{(-k)(-m)}$ are the training subsets, and $L_k$ the validation folds, for $k=1,...,M$. Let $\hat{f}_{Ej}^{(-k)(-m)}(\boldsymbol{x})= f^{(-k)(-m)}_{Ej} (\boldsymbol{x},\hat{\boldsymbol{\theta}})$ denote $j$-th level-1 learner estimated on training subset $L^{(-k)(-m)}$.

Given the partition $CV^{m}$, we have the formulation of two-step ensemble learning given by Equation (7).

We reproduce below Equation (8) for convenience, which represents the level-2 ensemble learner based on weight vector $\beta_{m}$, for predictions of a "new" data $x$ on hold-out dataset, given $CV^{m}$  as 

\[
\begin{split}
f_{E123^{m}}(\boldsymbol{x},\boldsymbol{\beta}) = \beta_{1}\hat{f}_{E1^{m}}(\boldsymbol{x}) + \beta_{2}\hat{f}_{E2^{m}}(\boldsymbol{x}) + \beta_{3}\hat{f}_{E3^{m}}(\boldsymbol{x}) \\
=  \beta_{1}f_{E1^{m}}(\boldsymbol{x},\hat{\boldsymbol{\alpha}}_1) + \beta_{2}f_{E2^{m}}(\boldsymbol{x},\hat{\boldsymbol{\alpha}}_2)+ \beta_{3}f_{E3^{m}}(\boldsymbol{x},\hat{\boldsymbol{\alpha}}_3)
\end{split}
\]

In order to summarize the method of two-step level-2 learning from top to bottom, let us emphasized that the weight vector $\boldsymbol{\beta}_{m}$ for $f_{E123^{m}}$ or, level-2 learner, present in Equation (8) is estimated based on training datasets of partition $CV^{m}$ using $level$-2 data as input to Equation (7):
predictions of level-1 learners $f^{(-k)(-m)}_{Ej}\allowbreak(\boldsymbol{x}_n,\hat{\boldsymbol{\alpha}}_j)$ along with actual values $y_n$, where $\boldsymbol{x}_n  \in L_k$, and  $L_k$ for $k=1,2,..., K$ are validation folds.. 

The optimal weight vector $\hat{\boldsymbol{\alpha}}_j$, for $j=1,2,3$, corresponding to a parameter for each level-1 learner approximation, $f^{(-k)(-m)}_{Ej}(\boldsymbol{x}_n,\hat{\boldsymbol{\alpha}}_j)$ with $j=1,2,3$ in Equation (7), was estimated previously by solving an optimization problem similar to Equation (3), but in the context of a $K$-fold $CV^{m}$ partition, using level-1 data as input to such optimization:  predictions of level-0 learners $f^{(-k)(-m)}_{i_{j}}(\boldsymbol{x}_n,\hat{\boldsymbol{\theta}})$, along with actual values $y_n$, where $\boldsymbol{x}_n  \in L_k$, and  $L_k$ for $k=1,2,..., K$ are validation folds.. For $j=1$ the index $i_1$ takes values in set $\{1,2,3\}$ corresponding to base learners in $f_{E1}$. On the other hand, for $j=2$, index $i_2$ takes values across set $\{1,2,4\}$ corresponding to base learners in $f_{E2}$, and for $j=3$, index $i_3$ takes values in set $\{1,3,4\}$ corresponding to base learners in $f_{E3}$.

\subsection{Experimental Set-up}
\label{sec:6.2}
In order to evaluate empirically max-min systematic two-step level-2 ensemble learning, $f_{E{123^*}}$, we consider four simple base models already introduced in Section 3.1: constant regression, $f_1$, linear regression, $f_2$, quadratic regression, $f_3$ and radial basis networks, $f_4$, and take the best model in terms of error performance for each database to build an oracle, which we called $Best$. We also consider three standard stacking approaches, or level-1 learners, $f_{E1}$, a linear combination of base learners in set $\{f_1, f_2, f_3\}$, $f_{E2}$, a linear combination of base learners in set $\{f_1, f_2, f_4\}$, and $f_{E3}$, a linear combination of base learners in  $\{f_1, f_3, f_4\}$, and compare $Best$ versus the best performer, $f_{E^*}$. We are also be comparing $f_{E{123^*}}$ versus $f_{E^*}$, and against $Best$. 

The outputs, or predictions of our base models $f_1$, $f_2$, $f_3$, and $f_4$ are represented in the next tables, and figures, by the acronyms $CR$, $LR$, $QR$ and $RBF$ respectively. These acronyms represent common simple models, which were already introduced in Section 3.1. So, $f_{E1}$ is a combination of  $\{CR, LR, QR\}$, $f_{E2}$ is a combination of $\{CR, LR, RBF\}$, and $f_{E3}$ is a combination of $\{CR, QR, RBF\}$.

In addition to that, we also compared our results versus $\textit{GLMNET}$ package \citep{friedman10} implemented in $R$, a statistical programming software tool \citep{r2008}. We also use two regression tree ensemble algorithms which are provided in $\textit{Weka}$ \citep{witten05data} \citep{hall2009weka}, a data mining software tool. We use the $\textit{RWeka}$ \citep{rweka} interface to call these two regression tree algorithms from $R$. One is known as $M5P$ and the other $Bagging$$-$$M5P$ \citep{quinlan92}, \citep{wang97}. We use the $\textit{Weka}$ default parameters for $M5P$, and $Bagging$$-$$M5P$, and the $GLMNET$ default parameters for the generation of their corresponding predictions. 

The empirical study uses thirty databases which are popular in the literature, most of them from the UCI Repository of Machine Learning Databases and Domain Theories \citep{blake98}. For each data set, we split it randomly in two parts, where the first part has $80$ percent of the observations used for training, and the remaining $20$ percent of the observations is used for testing. We conduct a $K$-fold $CV$ partition with $K=5$ on the training part to fit the models in our approach, and measure the performance on the testing part. For the base models of the oracle $Best$, and the three state-of-the art ensemble algorithms against we are comparing our results, we pass the whole training part to each method to fit its parameters, and then we measure their performance on the test part for comparison purposes. Then, a non-parametric two-sided Wilcoxon signed rank test \citep{wilcoxon45} is applied, as suggested in \citet{demsar06}, to compare two approaches over all data sets with a significance level of 0.05. The null hypothesis is that the average difference between both approaches is zero. The $p$-value is compared against the $0.05$ confidence level to determine the null hypothesis must be accepted or rejected. Alternatively, we compute the $95\%$ confidence interval around a point estimate to decide whether the zero average difference falls within the interval to accept or reject the null hypothesis.  

\section{Experimental Results}
\label{sec:7}
We present in Section 7.1 the results regarding standard stacking level-1 learning, and in Section 7.2 the second extension to standard stacking approach: systematic level-2 learning. While in Section 7.3, we compare systematic level-2 ensemble learning against three of the state of the art ensemble learning algorithms for regression. 
 
\subsection{Results for level-1 Learning: Standard Stacking}
\label{sec:7.1}
In this section, we present different results, among them the comparison of each of the level-1 learners versus an oracle of databases composed of the best level-0 learner for each database. In addition to that, we select the best performer among the level-1 learners versus the database oracle.
In Table 1, please see Appendix A for all table results,  we present the $NMSE$ for the thirty databases considered in this work. 
If we look at Table 1, in the first column we have the database names. Next, we have split such table in two parts, in the first half we have the $NMSE's$ for the level-0 learners:  ${CR}$, ${LR}$, ${QR}$ and $RBF$. On the other hand, in the second half of the table, we have the $NMSE's$ for the level-1 ensemble $f_{E1}$, linear combination of base learners in $\{CR, LR, QR\}$, followed by  the other two level-1 ensembles $f_{E2}$, linear combination of base learners in $\{CR, LR, RBF\}$, and $f_{E3}$, linear combination of base learners in $\{CR, QR, RBF\}$. The ensemble weights are determined using the optimal way as we described in section 3.2. Finally, in the last two columns are the number of records, and the total number of variables included in each database, independent variables plus dependent variable. It should be mentioned that we will build a database oracle based on the best level-0 model performer in each database, which we will compare later, in terms of error performance, against the best level-1 learner, as we will describe shortly.

We will first focus our analysis on the level-0 learners, first half of the table.  
We can observe from Table 1 that the prediction performance of the level-0 learner $LR$, as measured by the $NMSE$, is the best, indicated as underlined bold error, in ten out of thirty databases among the four models, $CR$, $LR$, $QR$ and $RBF$. The model ${QR}$ is the best in thirteen out of the thirty databases. While the model $RBF$ is the best in seven of such databases. Here, we build $Best$ the oracle which consists of the best level-0 model performer in terms of the $NMSE$ for each database. The average error performances across all databases, in terms of $NMSE's$, are $1.0079$, $0.5746$, $2.1527$, $0.6136$ for $CR$, $LR$, $QR$ and $RBF$ level-0 learners respectively. Among the level-0 learners the $LR$ model has the best error performance with a $0.5746$ $NMSE$ across all databases. 

In the second part of Table 1, the level-1 learner $f_{E1}$ is the best, among the level-1 learners, in ten out of thirty databases, while $f_{E2}$ is the best in seven of those databases, and $f_{E3}$ is the best in eleven databases. In addition to that, there are two ties, where for $Concrete$ and $Solar Flares$, the $NMSE's$ of $f_{E1}$ are tied with the corresponding errors for $f_{E3}$ and $f_{E2}$ respectively. In relation to the level-1 learners their average error performances across all databases, are $0.5331$, $0.5481$, and $0.5418$ for $f_{E1}$, $f_{E2}$ and $f_{E3}$ respectively. Then, the best level-0 learner performer is $f_{E1}$ with a $NMSE$ of $0.5331$. Next, we will compare the performance results between the database oracle and each of the level-1 learners to determine which level-1 learner performs better in average versus the oracle.

In Fig. 1, we visualize the difference of the $NMSE$'s between the oracle, $Best$, or best performer among the models $CR$, $LR$, $QR$ and $RBF$ on the testing part, and the ensemble $f_{E1}$ for each of the thirty databases considered. The differences are sorted by its absolute values. A positive difference, right part of the figure, means that accuracy is in favour of $f_{E1}$, while a negative difference means that the accuracy is against it. We will use this convention for all figures where we present the differences in performance between models or ensembles. By looking at the positive values in Fig. 1, we see that the ensemble $f_{E1}$ has better error performance than the oracle in the following databases: $Wisonsin$, $Sensor$, $ Cpu$, $Strike$, $Pollut$, $BetDiet$, $SolarFlares$, $RetDiet$, $HouseB$, $PM10$, $Imports$, $Wages$ and $AutoMpg$. It is worth to mention that the best performer in each database is not know in advance, as we are using the hold-out sample as testing dataset.

The result of two-sided non-parametric Wilcoxon signed-rank test to evaluate the performance of the oracle $Best$, and the ensemble $f_{E1}$  gives a $p$-value of $0.033$.  We reject $Ho$, the null hypothesis that there was not significant difference in accuracy, in terms of $NMSE$, between the database oracle $Best$, and the ensemble $f_{E1}$, i.e. $Ho$ stated that the average difference between their corresponding $NMSE$s is zero. The $95\%$ confidence interval is $(-0.0514,-0.00123)$, with a point estimate of the average difference is $-0.017$, which lead us to conclude a better performance of $Best$ vs $f_{E1}$ as the zero average difference is to the right of such confidence interval.

\begin{figure}[h]
\centering
\includegraphics[height=8.5cm,width=13cm]{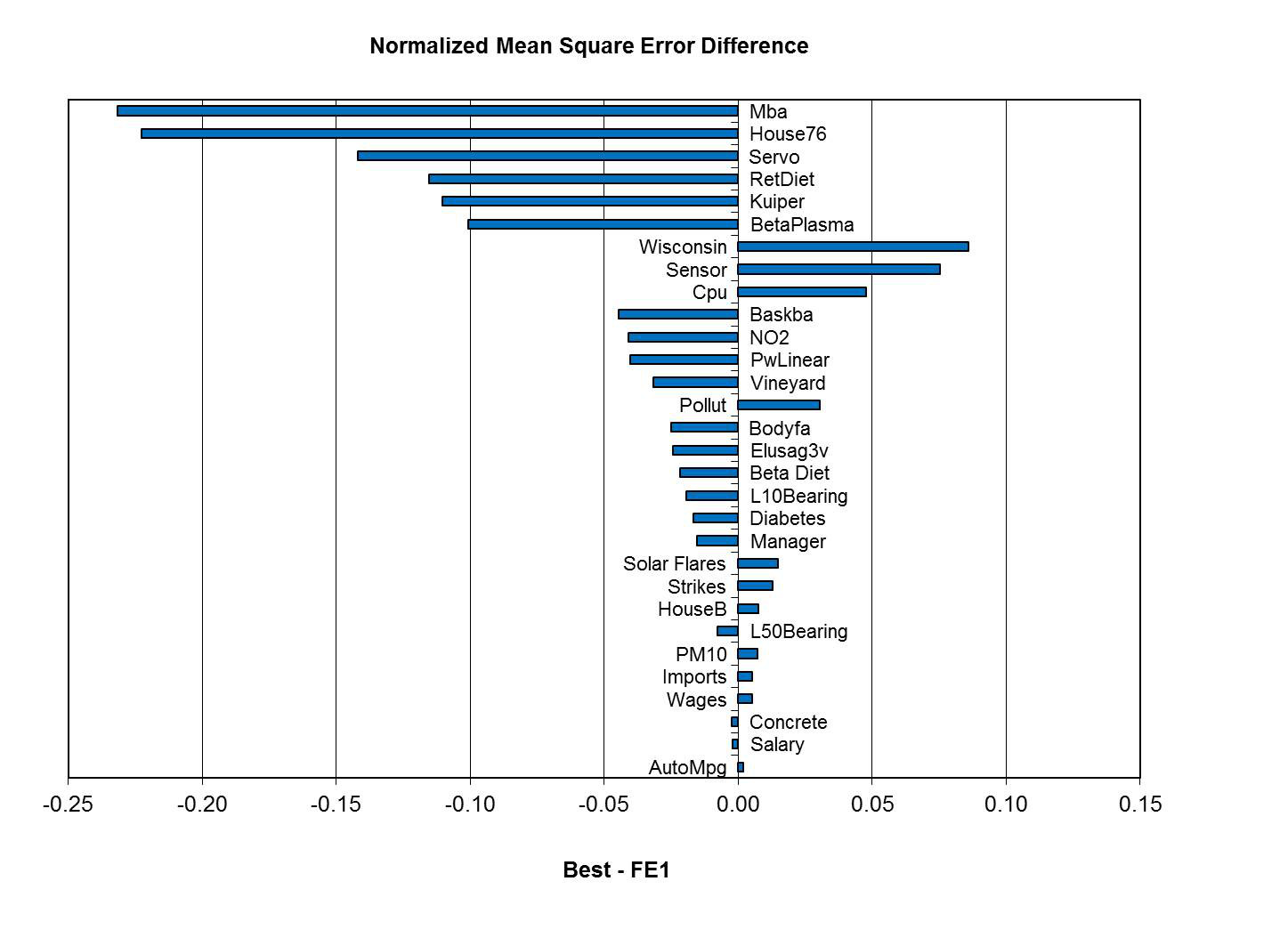}
\end{figure}
\begin{center}
\textbf{Fig. 1} Normalized Mean Square Error Difference \\
 Results of \{$Best$\} vs  \{$f_{E1}$\}   
\end{center}

On the other hand, for the ensembles $f_{E2}$ and $f_{E3}$ we do not present the result figures of each of them versus the oracle $Best$ , or best performer among the models $CR$, $LR$, $QR$ and $RBF$ on the testing part across the thirty databases considered. However, the result of two-sided non-parametric Wilcoxon signed-rank tests to evaluate the performance of the oracle $Best$, and the ensemble $f_{E2}$  gives a $p$-value of $0.0011$. The $95\%$ confidence interval is $(-0.0654,-0.0132)$ with a point estimate of the average difference of $-0.036$. We reject the null hypothesis Ho, and therefore $Best$ outperforms $f_{E2}$ in prediction accuracy.

The result of two-sided non-parametric Wilcoxon signed-rank tests to evaluate the performance of the oracle $Best$, and the ensemble $f_{E3}$  gives a $p$-value of $0.0021$. The $95\%$ confidence interval is $(-0.0437,-0.0085)$, with a point estimate of the average difference of $-0.0238$. We reject Ho, and $Best$ outperforms $f_{E3}$ in prediction accuracy, as the average difference between both models being different than zero is significant. On the other hand, the level-1 learner which performed better, in terms of $p$-value, against the database oracle $Best$ is $f_{E1}$, so we select it as the best level-1 learner, and it will be denoted as $f_{E^*}$ from now on. 

\subsection{Results for Max-Min Rule-based Systematic level-2 Learning: Second Extension to Standard Stacking}
\label{sec:7.2}
In Section 7.2.1, we first present the results of applying the Algorithm $S$ from Section 5.2.4 to select the best max-min rule-based level-2 learner, or corresponding partition with the aim of improving the predictions from the original level-2 learner. In Section 7.2.2, we compare the performance results of this best level-2 learner against the best level-1 learner, then in Section 7.2.3, we compare the performance of best level-2 learner against the oracle of best models. 

\subsubsection{Best Systematic level-2 Learner}
\label{sec:7.2.1}
In order to improve the two-step level-2 learning, we have introduced the methodology of  systematic two-step level-2 learning for the mixing of level-1 learners. We will illustrate Algorithm $S$ from Section 5.2.4 by using it to select the $CV$ partition, and applying it on each of the thirty databases considered. We will present the corresponding $NMSE$'s of systematic two-step level-2 learning $f^{}_{E123^{m}}$,  for $m=1,2,...,K$, with $K=5$. 

In Table 2, we show the corresponding results, where in the first column is as usual the database name, followed by the next six columns corresponding to the Normalized Mean Square Errors $(NMSE's)$ of the ensembles $f^{}_{E123^{0}}$, $f^{}_{E123^{1}}$, $f^{}_{E123^{2}}$, $f^{}_{E123^{3}}$, $f^{}_{E123^{4}}$ and $f^{}_{E123^{5}}$. The next column $Rule$ refers to the selection rule that was employed to select the $CV$ partition, among $CV^0$, $CV^{1}$, $CV^{2}$, $CV^{3}$, $CV^{4}$ and $CV^{5}$, on which we are going to make the final prediction, partition $CV^{*}$. For such selection we used the two max-min rule-based based criteria described in Algorithm $S$. The selected $CV^{*}$ is indicated as underlined error in Table 2 for each of the databases considered.

If we look again at Table 2, where we have the original two-step level-2 learner,  $f_{E123^0}$,  versus  systematic two-step level-2 learners, $f_{E123^{m}};m=1,...,5$, we see that we have improved,  indicated as bold errors, in fifteen databases: $MBA$, $Manager$, $CPU$, $Wisconsin$, $L50Bearing$, $BodyFat$, $RetDiet$, $BetaPlasma$. $BetaDiet$, \textit{AutoMpg}, $HouseB$, $Sensor$, $PM10$, $Wages$ and $Solar Flares$. The corresponding selected best max-min level-2 learners for such databases by the Algorithm $S$ are $f_{E123^{5}}$, $f_{E123^{3}}$, $f_{E123^{5}}$,$f_{E123^{1}}$, $f_{E123^{1}}$, $f_{E123^{3}}$, $f_{E123^{4}}$, $f_{E123^{3}}$, $f_{E123^{3}}$, $f_{E123^{1}}$, $f_{E123^{1}}$, $f_{E123^{3}}$, $f_{E123^{1}}$,  $f_{E123^{2}}$ and  $f_{E123^{3}}$, respectively. On the other hand, for nine databases the Algorithm $S$ selected $f_{E123}$, which is exactly the original level-2 learner, $f_{E123}$. In the remaining six databases we lost a bit of accuracy where, e.g., in $Elusag3v$, the algorithm selected max-min level-2 learner $f_{E123^{(4)}}$ with an error of $.302$, the other five databases are $Servo$, $PWLinear$, $NO2$, $Strikes$ and $Kuiper$. 

In Table 3, we present the $NMSE$'s of $f^{}_{E123^{*}}$ corresponding to the best max-min  level-2 learners selected by Algorithm $S$, which are represented in Table 2 as underlined errors. Now, we compare them against the corresponding $NMSE$'s of $f_{E*}$, the best level-1 learner. In order to mark in Table 3 the ensembles with best error performance between the best level-1 learning and systematic level-2 learning for each database, we highlight theirs corresponding $NMSE$'s in bold.  

As we can see from Table 3, the $NMSE$'s of $f_{E123^{*}}$ are the best, indicated as bold errors under its corresponding column, against the corresponding $NMSE$ of $f_{E^*}$, in nineteen out of thirty databases. The best level-1 learner $f_{E^*}$ is the best in eight databases. While for three databases the error performance is the same between them.

\subsubsection{Best level-2 Learner against Best level-1 Learner}
\label{sec:7.2.2}
In Fig. 2, we are visualizing the $NMSE$'s differences of  max-min systematic level-2 learning, and level-1 learning for the thirty databases considered.

We can clearly see that the best max-min level-2 learner  $f_{E123^*}$ is better than the best level-1 learner $f_{E^*}$ in the following nineteen databases: $House76$, $Kuiper$, $Servo$, $RetDiet$, $BetaPlasma$,  $PM10$, $Manager$, $Wages$,  $BodyFa$, $BetaDiet$,  $MBA$, $NO2$, $PWLinear$, $Strikes$, $HouseB$, $Sensor$, $CPU$, $AutoMpg$, and\textit{ SolarFlares}. The result of $95\%$ two-sided non-parametric Wilcoxon signed-rank test to evaluate the performance of the best level-1 ensemble, $f_{E^*}$, and the ensemble $f_{E123^*}$  gives a test statistics $t = 1.754$ with $p$-value of $0.079$. So, we do not reject Ho. The $95\%$ confidence interval is $(-0.00011,0.023)$, and the point estimate of the average difference is $0.0078$. However, the $90\%$ two-sided non-parametric Wilcoxon signed-rank test is significant to reject $Ho$ with a $p$-value of $0.079$. The $90\%$ confidence interval is (0.00083, 0.020) with a point estimate of $0.0078$, where the zero average difference corresponding to the null hypothesis is outside and to the left of such interval. Therefore, the ensemble $f_{E123^*}$ outperforms $f_{E^*}$ at a $90\%$ significance level. 
\begin{figure}[h]
\centering
\includegraphics[height=8.3cm,width=13cm]{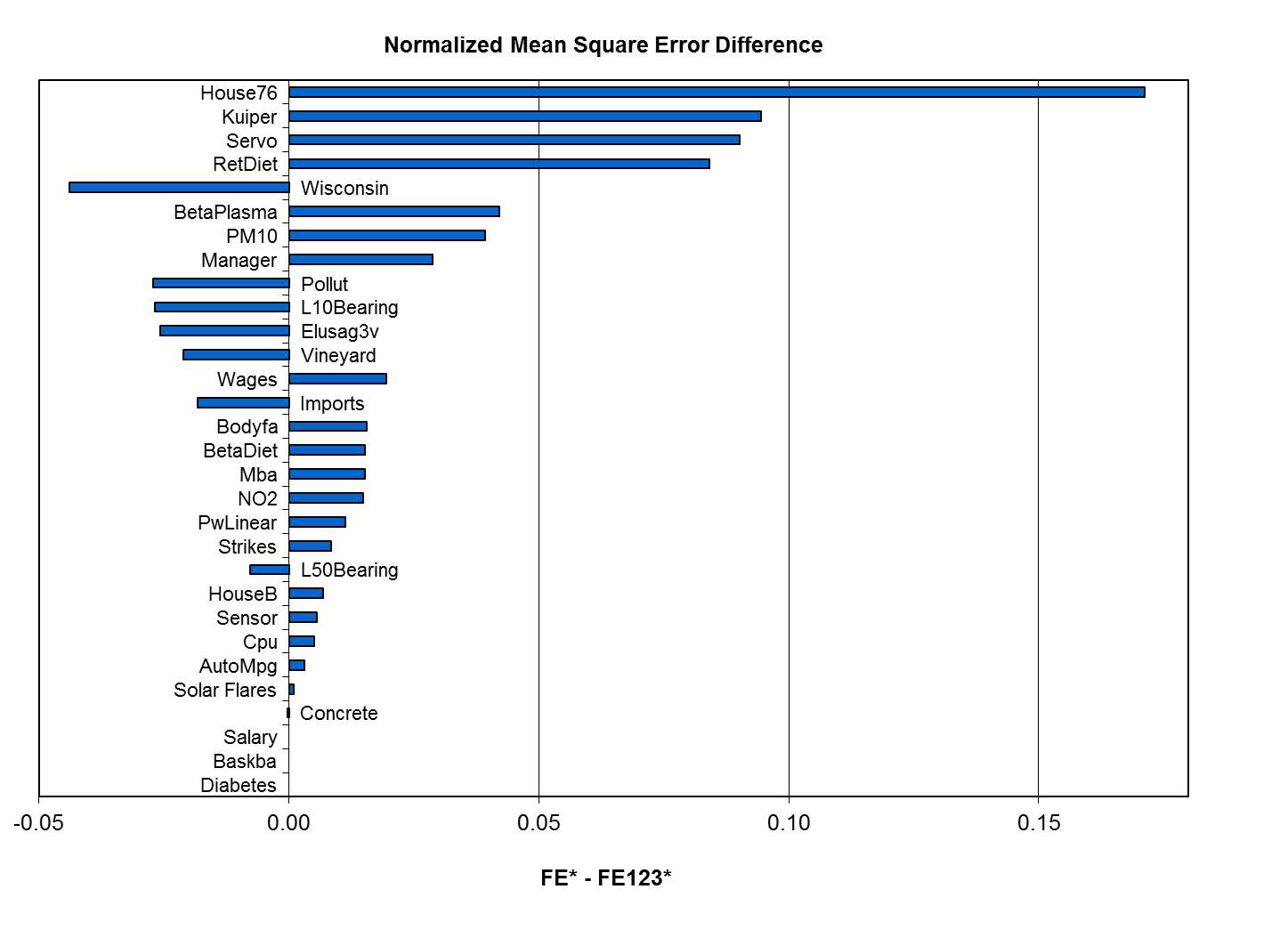}
\end{figure}
\begin{center}
\textbf{Fig. 2} Normalized Mean Square Error Difference \\
 Results of \{$f_{E^*}$\}   vs \{$f_{E123^{*}}$\}      
\end{center}

Based on the above results, Table 3 and Fig. 2, we see that max-min systematic level-2 learning with  $f_{E123^*}$ is a step in the right direction being in average more accurate and robust than the best level-1 learning $f_{E^*}$. Therefore, we are achieving our goal of improving the performance error of best level-1 ensemble, $f_{E^*}$, through the two-step level-2 learning and systematization proposed in this work.

\subsubsection{Best level-2 Learner against Oracle Best}
\label{sec:7.2.3}
In Fig. 3, we present the $NMSE$'s difference between the best performer, or database oracle, among the models $CR$, $LR$, $QR$ and $RBF$ on the testing part, and the best max-min level-2 learner $f_{E123^*}$ for each of the thirty databases considered. By looking at the positive value part of such figure, we see that $f_{E123^*}$ has better error performance than the database oracle, $Best$, in eleven out of thirty databases: $Sensor$, $Cpu$, $PM10$, $Wisconsin$,  $Wages$, $Strikes$, $SolarFlares$, $HouseB$, $Manager$, $AutoMpg$ and $Pollut$. 
The result of $95\%$ two-sided non-parametric Wilcoxon signed-rank test to evaluate the performance of the oracle $f_{Best}$, and the ensemble $f_{E123^*}$  gives a $p$-value of $0.147$. So, we do not reject $Ho$, as there is not significant accuracy difference between the oracle $Best$, and the two-step systematic ensemble learning $f_{E123^*}$. The $95\%$ confidence interval is $(-0.026,0.0035)$, and the point estimate of the average difference is $-0.011$.
\begin{figure}[h]
\centering
\includegraphics[height=8.5cm,width=13cm]{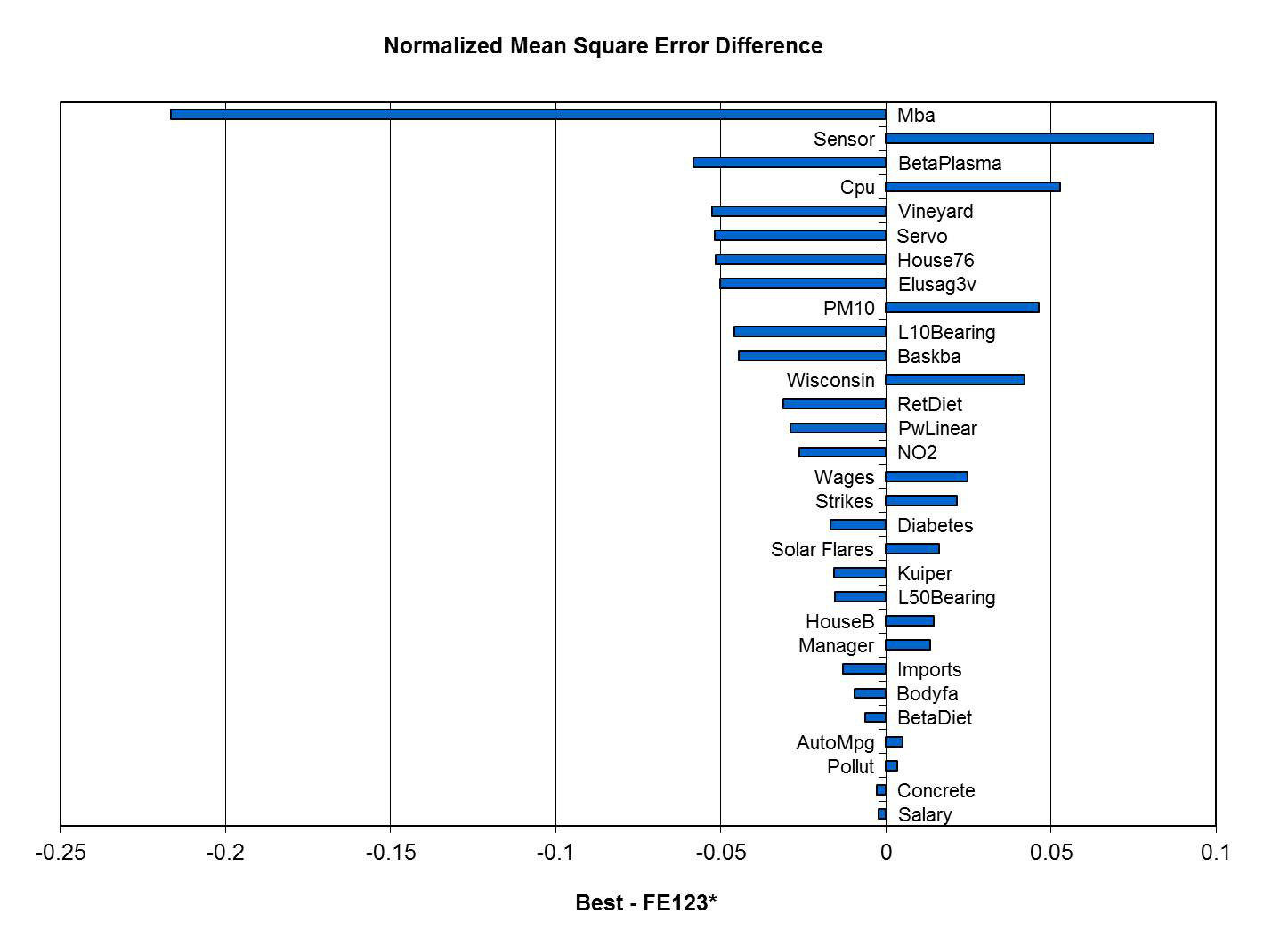}
\end{figure}
\begin{center}
\textbf{Fig. 3} Normalized Mean Square Error Difference \\
 Results of \{$Best$\}  vs \{$f_{E123^{*}}$\}  
\end{center}

\subsection{Best level-2 Learner against State of the Art Ensemble Methods}
\label{sec:7.3}
For the three state-of-the art ensemble algorithms, $\textit{GLMNET}$, $M5P$, and $Bagging-M5P$, we pass to each of them the same  training and test datasets we used in our methods, to fit its parameters, and to measure their performance for comparison purposes versus our results in this paper.

The above three methods are well regarded state-of-the-art ensemble learning algorithms for regression problems.
The two-step level-2 systematic ensemble learning presented in this paper constitutes a new state-of-the-art ensemble learning algorithm, as we have tested it against each of the three ensemble learning algorithms just mentioned, and we have gotten excellent results. Our ensemble learning algorithm performed better than $\textit{GLMNET}$ and $\textit{M5P}$  in terms of error performance, and it is as good as $\textit{Bagging$$-$$M5P}$. In this section we present results that support our above claims. 
     
\subsubsection{Best level-2 Learner against \textit{GLMNET}}
\label{sec:7.3.1}
The $\textit{GLMNET}$ package \citep{friedman10}, $(www.jstatsoft.org/v33/i01)$ , is available in the $R$ statistical software \citep{r2013}, $(http://www.R-project.org)$. $\textit{GLMNET}$ solves the penalized residual sum of square, that can perform lasso penalized regression, ridge regression, and the elastic net. We use the version 3.01 of $R$ software, and version 1.9-3 of $\textit{GLMNET}$ package to get the results presented in this paper. The lasso penalty encourages sparsity, so this penalty performs a sort of variable selection as some of the coefficients are shrunken to zero, while the ridge penalty encourages highly correlated variables to have similar coefficients. The elastic net is a linear combination of the two penalties  through an $\alpha$  parameter, with $0 \leq \alpha \leq 1$. Note that elastic net becomes the lasso when $\alpha = 1$, and the ridge regression when $\alpha = 0$. It has also a tuning parameter $\lambda \ge 0$, which is a complexity parameter. In our experiments, we used the default parameter for $\alpha$, which is $\alpha = 1$. For the $\lambda$ parameter we estimated it using cross-validation with its default parameter, and we take the minimum of $\lambda$ values as an estimate. 

In Fig. 4, we present the $NMSE$'s difference on the testing part between the $\textit{GLMNET}$ method, and the best max-min level-2 learner $f_{E123^*}$ for each of the thirty databases considered.

\begin{figure}[h]
\centering
\includegraphics[height=8.5cm,width=13cm]{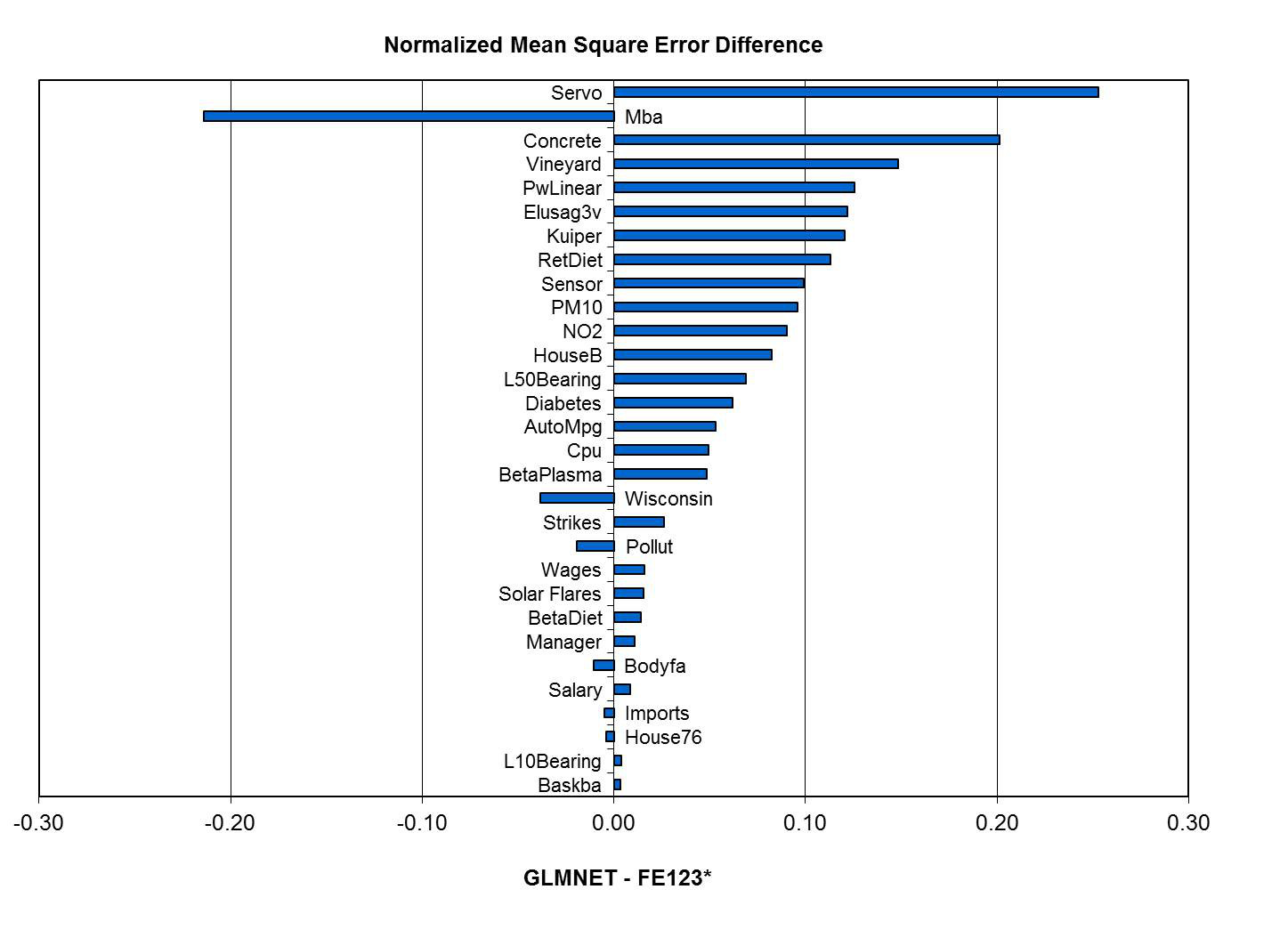}
\end{figure}
\begin{center}
\textbf{Fig. 4} Normalized Mean Square Error Difference \\
 Results of \{$GLMNET$\}   vs   \{$f_{E123^{*}}$\}   
\end{center}

The result of $95\%$ two-sided non-parametric Wilcoxon signed-rank test to evaluate the performance of $\textit{GLMNET}$, and the ensemble $f_{E123*}$  gives a $p$-value of $0.001$. So, we reject $Ho$ in favour of  $f_{E123^*}$, as there is a significant accuracy difference between $\textit{GLMNET}$, and the two-step systematic ensemble learning $f_{E123^*}$. The $95\%$ confidence interval is $(0.024,0.078)$, and the point estimate of the average difference is $0.051$.

\subsubsection{Best level-2 Learner against $M5P$}
\label{sec:7.3.2}
We use a regression tree algorithm $\textit{M5P}$ \citep{wang97}, which is based on an earlier $M5$ regression tree \citep{quinlan92}. $M5P$ algorithm is available in $\textit{Weka}$ data mining software \citep{witten05data} \citep{hall2009weka}, $(http://www.cs.waikato.ac.nz/\allowbreak ml/weka/)$. Although we install $\textit{Weka}$ version 3.6.1, we will only need to interact with it from $R$. We use the $\textit{RWeka}$ \citep{rweka} interface version 0.4-13 to call this regression tree algorithm from $R$. We use the $\textit{RWeka}$ default parameters for $\textit{M5P}$ for the generation of their corresponding predictions. 

$\textit{M5P}$ is a regression tree algorithm, which produces a regression tree such that each leaf node consists of a linear model for combining the numerical features. $\textit{M5P}$ uses a tree pruning mechanism, which is a sort of variable selection, and $\textit{M5P}$ can perform non-linear regression with the partitions provided by the internal nodes and is thus more powerful than linear regression. 

In Fig. 5, we present the $NMSE$'s difference on the testing part between $M5P$ method, and the best max-min level-2 learner $f_{E123*}$ for each of the thirty databases considered. 

\begin{figure}[h]
\centering
\includegraphics[height=8.5cm,width=13cm]{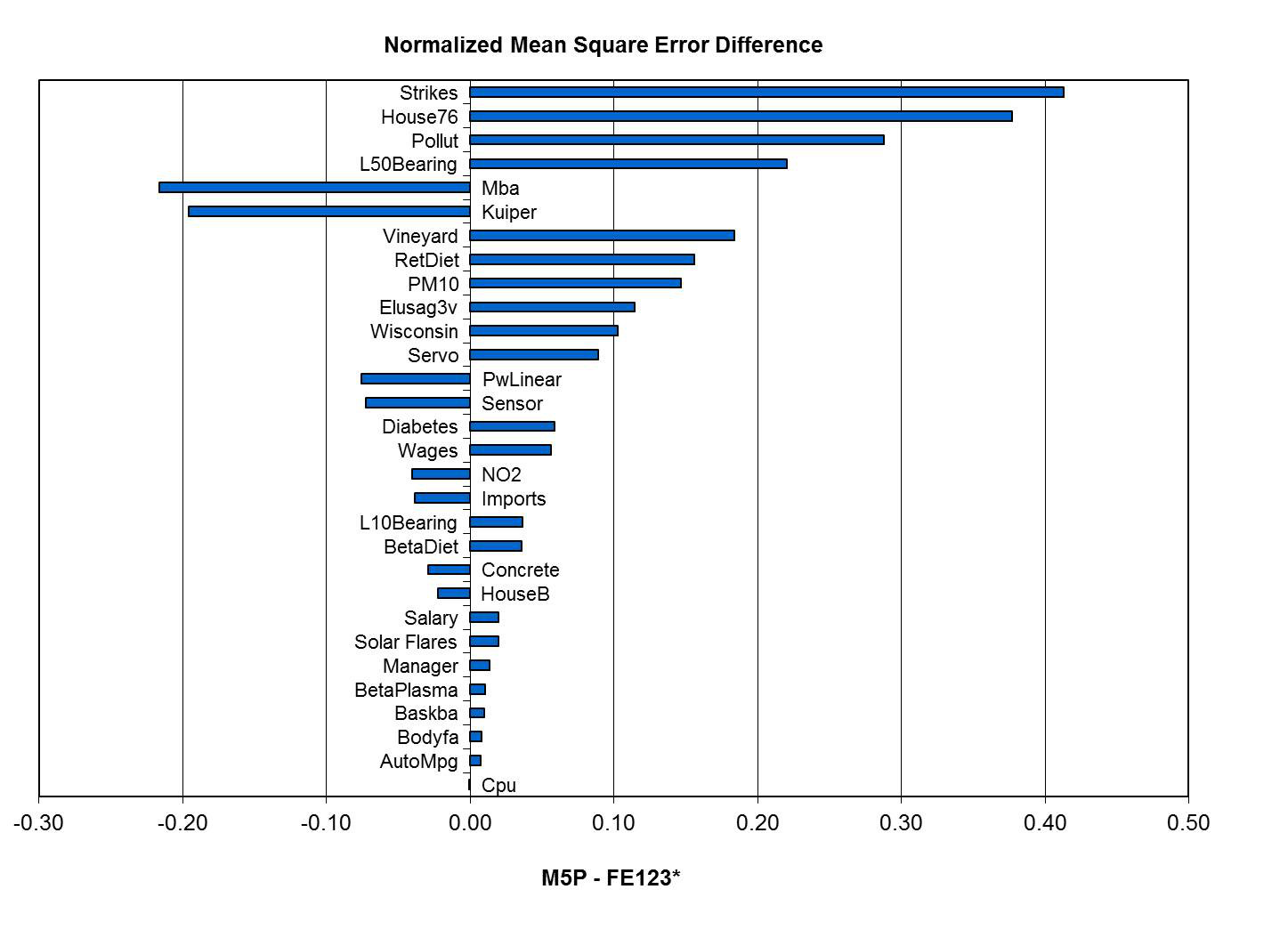}
\end{figure}
\begin{center}
\textbf{Fig. 5} Normalized Mean Square Error Difference \\
 Results of  \{$M5P$\}   vs   \{$f_{E123^{*}}$\}   
\end{center}

The result of $95\%$ two-sided non-parametric Wilcoxon signed-rank test to evaluate the performance of $\textit{M5P}$, and the ensemble $f_{E123*}$  gives a $p$-value of $0.041$. So, we reject $Ho$ in favour of $f_{E123*}$, as there is significant accuracy difference between $M5P$, and the two-step systematic ensemble learning $f_{E123*}$. The $95\%$ confidence interval is $(0.0034,0.096)$, and the point estimate of the average difference is $0.038$.

\subsubsection{Best level-2 Learner against \textit{Bagging-M5P}}
\label{sec:7.3.3}
In this part, we consider multiple $\textit{M5P}$ trees combined by the popular bootstrap aggregation (bagging) method (Breiman 1996), which is available in $\textit{Weka}$, and is called $\textit{Bagging-M5P}$ \citep{wang97}. We use again the $\textit{RWeka}$ \citep{rweka} interface to call this regression tree algorithm from $R$. We use the $\textit{RWeka}$ default parameters for $\textit{Bagging-M5P}$ for the generation of their corresponding predictions. 

In Fig. 6, we present the $NMSE$'s difference on the testing part between $Bagging-M5P$ method, and the best max-min level-2 learner $f_{E123*}$ for each of the thirty databases considered. 

The result of $95\%$ two-sided non-parametric Wilcoxon signed-rank test to evaluate the performance of $M5P$, and the ensemble $f_{E123*}$  gives a $p$-value of $0.192$. So, we do not reject $Ho$, as there is not significant accuracy difference between $\textit{Bagging-M5P}$, and the two-step systematic ensemble learning $f_{E123*}$. The $95\%$ confidence interval is $(-0.012,0.051)$, and the point estimate of the average difference is $0.0218$.

\begin{figure}[h]
\centering
\includegraphics[height=8.5cm,width=13cm]{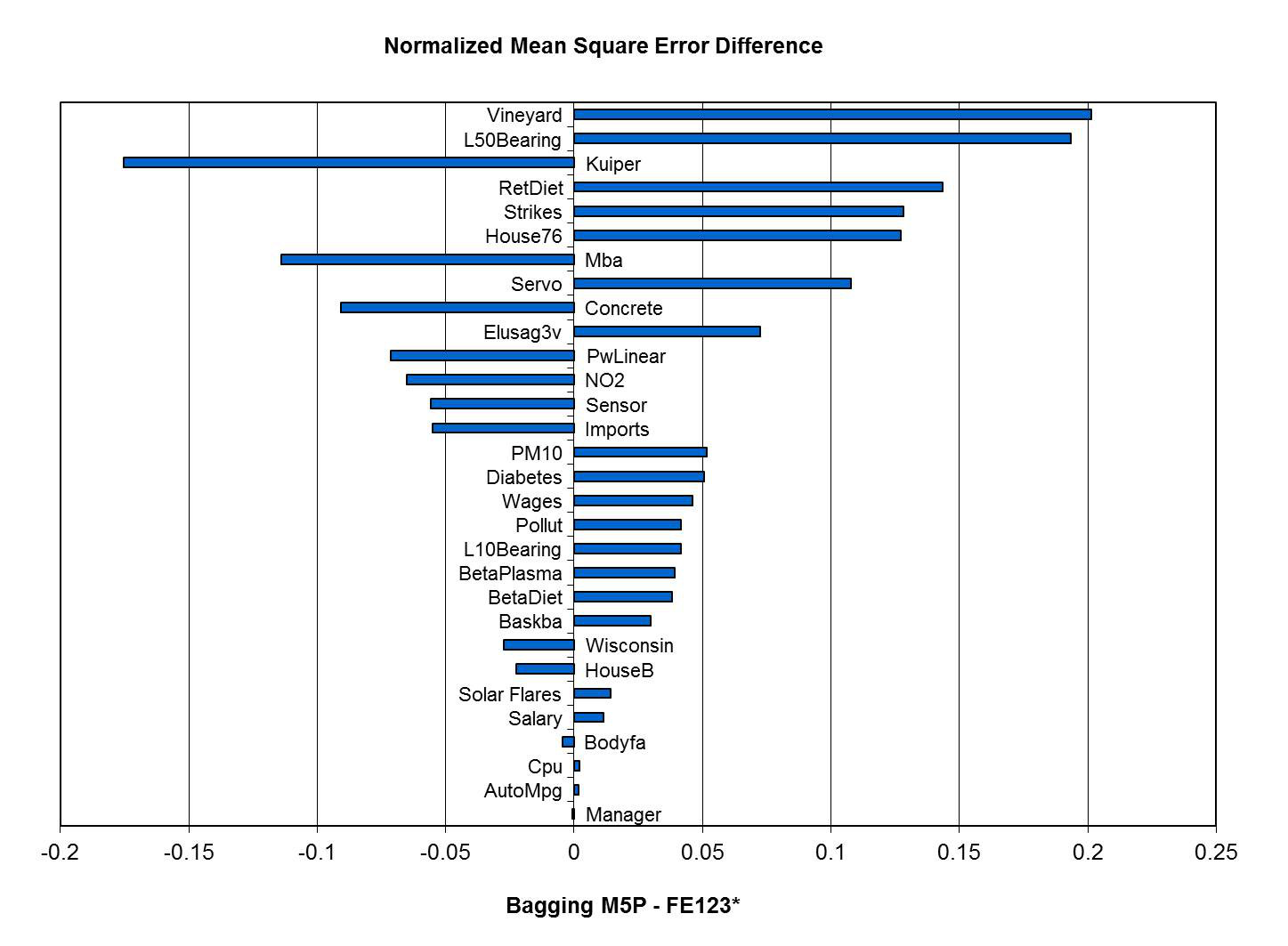}
\end{figure}
\begin{center}
\textbf{Fig. 6} Normalized Mean Square Error Difference \\
 Results of \{$\textit{Bagging-M5P}$\}   vs   \{$f_{E123^{*}}$\}    
\end{center}

\subsection{Discussion of Experimental Results}
\label{sec:7.4}
It should be emphasized that the iterative nature of two-step level-2 learning and its richness in structure in terms of correlation between each pair of ensembles was exploited in Section 5.1, through the generation of $CV$ partitions, and in Section 5.2, to select a partition on which to make the final prediction. 

A difference to what is developed for $GLMNET$ \citep{friedman10}, and the default parameter $\alpha=1$ used in this work, where the diversity comes by using lasso penalized regression as variable selection mechanism, $M5P$ \citep{quinlan92} whose diversity comes by pruning a regression tree, and $Bagging$$-$$M5P$ \citep{wang97}, which in addition to pruning a regression tree, creates more diversity by generating different training sub-samples through bagging; the diversity in our approach, which uses combinations of standard stacking approaches,  comes through the generation of different partitions on which such ensembles are trained. 

In addition to that, our approach controls the diversity of the different level-2 ensembles generated, or corresponding partitions, through a computed correlation criterion built on top of two-step ensemble learning, which provides an extra level of smoothing for the base ensembles in each partition. 

\section{Conclusions}
\label{sec:8}
We have empirically tested  three standard stacking approaches, or level-1 ensembles, each composed of a combination of simple based models, for heterogeneous stacking for regression. We have shown that the best of such ensembles performed not as good as the best base model from the ensemble by cross-validation. We have proposed two extensions to standard stacking approach. The first is to create an ensemble composed of different combinations of such standard approaches, through a two-step level-2 ensemble learning. We mentioned that the richness of structure in such two-step level-2 ensemble learning provided the basis to compute the base ensemble pairwise correlations, which will help us on  improving the prediction accuracy in the second extension. The second extension was built to systematically generate different partitions from the current partition, and correspondingly two-step level-2 ensembles; that along a partition correlation-based criterion, or a partition ranking-based criterion within an heuristic algorithm allowed us to select the best 2-level ensemble, and show that it performed better than the best of the standard stacking approaches, and is as good as the best of the base models from the standard stacking approaches by cross-validation. We also compared our results of best systematic level-2 ensemble learning versus three state of the art ensemble learning algorithms for regression, and found that our results performed better than two of those ensemble methods for regression, and it is as good as the third of such algorithms in terms of error performance.   

As a future research to do, it is worth to mention, that the systematic way to generate $CV$ partitions introduced in Section 5.1.1 will allow us to rank them, and reduce the size of the original $CV$ partition by eliminating the least relevant fold, and apply iteratively max-min rule-based Algorithm $S$. We are currently investigating different stopping criterion that could prevent us from iterating the algorithm one more time, if after the current iteration it could detect that we may not get better prediction accuracy going into the next iteration after pruning the current partition. 

\newpage

\appendix
\section*{Appendix A.}
\label{app:tables}
\begin{table}[h]
\centering
\resizebox{14cm}{!} {
\begin{tabular}{|r||r|r|r|r||r|r|r||r|r|}
 	\hline
$DB$ & $CR$ & $LR$ & $QR$ & $RBF$ &  $f_{E1}$ &$f_{E2}$&$f_{E3}$& $Rec$ & $Var$ \\
\hline
Vineyard & 1.051 & 0.403 & \textbf{0.204} & 0.353 & \textbf{0.236} & 0.390 & 0.255 & 43 & 3\\
\hline
Diabetes & 1.227 & 0.390 & \textbf{0.315} & 0.408 & \textbf{0.332} & 0.516 & 0.432 & 43 & 3\\
\hline
Pollut & 1.159 & \textbf{0.516} & 9.063 & 0.797 & \textbf{0.486} & 0.513 & 0.828 & 50 & 16\\
\hline
Mba & 1.126 & \textbf{0.604} & 0.842 & 0.898 & 0.835 & \textbf{0.682}& 0.843 & 51 & 3\\
\hline
Elusag3v & 1.158 & 0.418 & \textbf{0.252} & 0.624 & 0.277 & 0.390 & \textbf{0.256} & 55 & 3\\
\hline
House76 & 1.010 & \textbf{0.443} & 2.358 & 0.454 & 0.665 & \textbf{0.494} & 0.516 & 76 & 14\\
\hline
Baskba & 1.009 & 0.639 & 0.629 & \textbf{0.585} & 0.630 & 0.644 & \textbf{0.571} & 78 & 5\\
\hline
Servo & 1.109 & 0.720 & 0.508 & \textbf{0.406} & 0.548 & \textbf{0.422} & 0.433 & 140 & 3\\
\hline
Manager & 1.004 & \textbf{0.688} & 0.745 & 0.727 & 0.703 & \textbf{0.688} & 0.730 & 150 & 3\\
\hline
Imports & 1.000 & \textbf{0.190} & 0.551 & 0.249 & \textbf{0.185} & 0.205 & 0.227 & 162 & 16\\
\hline
PwLinear & 1.005 & 0.315 & \textbf{0.158} & 0.242 & 0.199 & 0.267 & \textbf{0.165} & 168 & 11\\
\hline
Cpu & 1.016 & \textbf{0.119} & 0.142 & 0.308 & \textbf{0.071} & 0.077 & 0.180 & 178 & 7\\
\hline
Wisconsin & 1.008 & \textbf{1.002} & 24.818 & 1.085 & \textbf{0.916} & 0.968 & 1.017 & 194 & 34\\
\hline
L10Bearing & 1.002 & 0.524 & \textbf{0.475} & 0.838 & \textbf{0.494} & 0.556 & 0.511 & 210 & 4\\
\hline
L50Bearing & 1.002 & 0.376 & \textbf{0.293} & 0.652 & \textbf{0.301} & 0.398 & 0.319 & 210 & 4\\
\hline
Bodyfa & 1.004 & \textbf{0.303} & 12.958 & 0.441 & 0.328 & \textbf{0.313} & 0.453 & 211 & 14\\
\hline
Salary & 1.005 & 0.304 & \textbf{0.295} & 0.376 & 0.298 & 0.304 & \textbf{0.296} & 220 & 4\\
\hline
BetaDiet & 1.013 & 0.955 & 1.205 & \textbf{0.909} & 0.930 & 0.918 & \textbf{0.906} & 315 & 11\\
\hline
BetaPlasma & 1.000 & 0.776 & 1.640 & \textbf{0.692} & 0.792 & 0.722 & \textbf{0.716} & 315 & 11\\
\hline
RetDiet & 1.008 & 1.046 & 0.963 & \textbf{0.834} & 0.949 & 0.895 & \textbf{0.834} & 315 & 11\\
\hline
AutoMpg & 1.000 & 0.216 & \textbf{0.166} & 0.202 & 0.163 & 0.183 & \textbf{0.161} & 329 & 8\\
\hline
HouseB & 1.013 & 0.216 & \textbf{0.149} & 0.255 & \textbf{0.142} & 0.211 & 0.163 & 425 & 14\\
\hline
Sensor & 1.000 & 1.054 & \textbf{0.999} & 1.050 & \textbf{0.924} & 1.016 & 0.929 & 483 & 12\\
\hline
NO2 & 1.003 & 0.549 & \textbf{0.432} & 0.526 & 0.473 & 0.481 & \textbf{0.433} & 500 & 8\\
\hline
PM10 & 1.009 & 0.847 & \textbf{0.781} & 0.795 & 0.774 & 0.771 & \textbf{0.752} & 500 & 8\\
\hline
Wages & 1.000 & \textbf{0.907} & 1.022 & 0.941 & 0.902 & \textbf{0.897} & 0.933 & 534 & 11\\
\hline
Strikes & 1.044 & 0.831 & 0.926 & \textbf{0.805} & 0.792 & 0.782 & \textbf{0.775} & 625 & 7\\
\hline
Kuiper & 1.022 & 0.530 & 0.506 & \textbf{0.395} & 0.505 & \textbf{0.397} & 0.408 & 804 & 8\\
\hline
Concrete & 0.203 & 0.400 & \textbf{0.200} & 0.522 & 0.203 & 0.400 & 0.203 & 865 & 9\\
\hline
Solar Flares & 1.024 & \textbf{0.957} & 0.985 & 1.040 & 0.942 & 0.942 & 1.011 & 1066 & 8\\

\hline

\end{tabular}  
}
\end{table}
\begin{center}
\textbf{Table 1} Normalized Mean Square Errors \\
level-0 Learners: \{$CR$\}, \{$LR$\},\{$QR$\}, \{$RBF$\} and level-1 Learners: \{$f_{E1}$\}, \{$f_{E2}$\}, \{$f_{E3}$\}     
\end{center}

\newpage

\begin{table}[h]
\centering
\resizebox{14cm}{!} {
\begin{tabular}{|r||r|r|r|r|r|r|r||r|r|}
 	\hline
$DB$ & $f_{E123^{0}}$ & $f_{E123^{1}}$ & $f_{E123^{2}}$ & $f_{E123^{3}}$ & $f_{E123^{4}}$ & $f_{E123^{5}}$ &$Rule$ & $Rec.$ & $Var.$ \\
\hline
Vineyard & \underline{0.257} & 0.306 & 0.344 & 0.404 & 0.398 & 0.238 & AR-L2 & 43 & 3\\
\hline
Diabetes & \underline{0.332} & 0.393 & 0.367 & 0.333 & 0.553 & 0.511 & NA & 43 & 3\\
\hline
Pollut & \underline{0.513} & 0.551 & 0.665 & 0.538 & 0.749 & 0.488 & AR-L2 & 50 & 16\\
\hline
Mba & 0.836 & 0.832 & 0.857 & 0.841 & .780 & \textbf{\underline{0.820}} & AR-L3 & 51 & 3\\
\hline
Elusag3v & 0.277 & 0.302 & 0.251 & 0.302 & \underline{0.302} & 0.303 & AR-L2 & 55 & 3\\
\hline
House76 & \underline{0.494} & 0.523 & 0.554 & 0.590 & 0.466 & 0.572 & AR-L2 & 76 & 14\\
\hline
Baskba & \underline{0.630} & 0.627 & 0.593 & 0.602 & 0.591 & 0.647 & NA & 78 & 5\\
\hline
Servo & 0.455 & 0.456 & 0.437 & 0.459 & \underline{0.458} & 0.475 & AR-L2 & 140 & 3\\
\hline
Manager & 0.701 & 0.694 & 0.710 & \textbf{\underline{0.674}} & 0.706 & 0.722 & AR-L2 & 150 & 3\\
\hline
Imports & \underline{0.203} & 0.211 & 0.202 & 0.210 & 0.192 & 0.155 & NA & 162 & 16\\
\hline
PwLinear & 0.187 & 0.201 & \underline{0.196} & 0.204 & 0.211 & 0.219 & MED & 168 & 11\\
\hline
Cpu & 0.077 & 0.119 & 0.060 & 0.138 & 0.093 & \textbf{\underline{0.066}} &  AR-L2 & 178 & 7\\
\hline
Wisconsin & 0.965 & \textbf{\underline{0.960}} & 0.992 & 0.938 & 0.964 & 0.904 & AR-L3 & 194 & 34\\
\hline
L10Bearing & \underline{0.521} & 0.522 & 0.549 & 0.506 & 0.498 & 0.501 &  AR-L2 & 210 & 4\\
\hline
L50Bearing & 0.320 &\textbf{\underline{ 0.308}} & 0.317 & 0.312 & 0.323 & 0.308 &  AR-L2 & 210 & 4\\
\hline
Bodyfa & 0.322 & 0.315 & 0.318 & \textbf{\underline{0.313}} & 0.359 & 0.336 &  AR-L3 & 211 & 14\\
\hline
Salary & \underline{0.298} & 0.304 & 0.301 & 0.298 & 0.299 & 0.299 & NA  & 220 & 4\\
\hline
RetDiet & 0.895 & 0.957 & 0.934 & 0.885 & \textbf{\underline{0.865}} & 0.873 & AR-L2  & 315 & 11\\
\hline
BetaPlasma & 0.751 & 0.687 & 0.732 & \textbf{\underline{0.750}} & 0.766 & 0.704 & AR-L2  & 315 & 11\\
\hline
BetaDiet & 0.918 & 0.906 & 0.922 & \textbf{\underline{0.915}} & 0.926 & 0.922 & MED & 315 & 11\\
\hline
AutoMpg & 0.160 & \textbf{\underline{0.160}} & 0.170 & 0.162 & 0.180 & 0.175 & AR-L2 & 329 & 8\\
\hline
HouseB & 0.151 & \textbf{\underline{0.135}} & 0.163 & 0.180 & 0.144 & 0.159 & MED & 425 & 14\\
\hline
Sensor & 0.929 & 0.927 & 0.924 & \textbf{\underline{0.918}} & 0.925 & 0.917 & MED & 483 & 12\\
\hline
PM10 & 0.753 & \textbf{\underline{0.735}} & 0.750 & 0.750 & 0.742 & 0.717 & AR-L2 & 500 & 8\\
\hline
NO2 & 0.451 & 0.467 & 0.472 & 0.488 & 0.453 & \underline{0.458} & MED & 500 & 8\\
\hline
Wages & 0.902 & 0.896 & \textbf{\underline{0.882}} & 0.887 & 0.869 & 0.881 & MED & 534 & 11\\
\hline
Strikes & 0.782 & 0.775 & \underline{0.783} & 0.784 & 0.788 & 0.781 & MED & 625 & 7\\
\hline
Kuiper & 0.408 & 0.408 & \underline{0.411} & 0.387 & 0.406 & 0.433 & AR-L2 & 804 & 8\\
\hline
Concrete & \underline{0.203} & 0.204 & 0.203 & 0.206 & 0.206 & 0.204 & MED & 865 & 9\\
\hline
Solar Flares & 0.942 & 0.947 & 0.944 & \textbf{\underline{ 0.941}} & 0.941 & 0.959 & AR-L2 & 1066 & 8\\
\hline

\end{tabular}  
}%end small
\end{table}
\begin{center}
\textbf{Table 2} Normalized Mean Square Errors \\
Systematic Generation of training sub-samples \\
First Iteration of algorithm \\
Results for \{$f_{E123^0}$\}, \{$f_{E123^{1}}$\},
 \{$f_{E123^{2}}$\}, \{$f_{E123^{3}}$\}, \{$f_{E123^{4}}$\}	and \{$f_{E123^{5}}$\}
\end{center}

\newpage

\begin{table}[h]
\centering
\small{
\begin{tabular}{|r||r|r||r|r|}
 	\hline
$DB$ & $f_{E*}$ & $f^{}_{E123^{*}}$ & $Rec$ & $Var$ \\

\hline
Vineyard & \textbf{0.236} & 0.257 & 43 & 3\\
\hline
Diabetes & 0.332 & 0.332 & 43 & 3\\
\hline
Pollut & \textbf{0.486} & 0.513 & 50 & 16\\
\hline
Mba & 0.835 & \textbf{0.820} & 51 & 3\\
\hline
Elusag3v & \textbf{0.277} & 0.302 & 55 & 3\\
\hline
House76 & 0.665 & \textbf{0.494} & 76 & 14\\
\hline
Baskba & 0.630 & 0.630 & 78 & 5\\
\hline
Servo & 0.548 & \textbf{0.458} & 140 & 3\\
\hline
Manager & 0.703 & \textbf{0.674} & 150 & 3\\
\hline
Imports & \textbf{0.185} & 0.203 & 162 & 16\\
\hline
PwLinear & 0.199 & \textbf{0.196} & 168 & 11\\
\hline
Cpu & 0.071 & \textbf{0.066} & 178 & 7\\
\hline
Wisconsin & \textbf{0.916} & 0.960 & 194 & 34\\
\hline
L10Bearing & \textbf{0.494} & 0.521 & 210 & 4\\
\hline
L50Bearing & \textbf{0.301} & 0.308 & 210 & 4\\
\hline
Bodyfa & 0.328 & \textbf{0.313} & 211 & 14\\
\hline
Salary & 0.298 & 0.298 & 220 & 4\\
\hline
RetDiet & 0.949 & \textbf{0.865} & 315 & 11\\
\hline
BetaPlasma & 0.792 & \textbf{0.750} & 315 & 11\\
\hline
BetaDiet & 0.930 & \textbf{0.915} & 315 & 11\\
\hline
AutoMpg & 0.163 & \textbf{0.160} & 329 & 8\\
\hline
HouseB & 0.142 & \textbf{0.135} & 425 & 14\\
\hline
Sensor & 0.924 & \textbf{0.918} & 483 & 12\\
\hline
PM10 & 0.774 & \textbf{0.735} & 500 & 8\\
\hline
NO2 & 0.473 & \textbf{0.458} & 500 & 8\\
\hline
Wages & 0.902 & \textbf{0.882} & 534 & 11\\
\hline
Strikes & 0.792 & \textbf{0.783} & 625 & 7\\
\hline
Kuiper & 0.505 & \textbf{0.411} & 804 & 8\\
\hline
Concrete &\textbf{ 0.203} & 0.203 & 865 & 9\\
\hline
Solar Flares & 0.942 & \textbf{0.941} & 1066 & 8\\
\hline
\end{tabular}  
}
\end{table}
\begin{center}
\textbf{Table 3} Normalized Mean Square Errors \\
Comparison of Best Max-Min level-2 Learner\\
versus Best level-1 Learner:\\
\{$f^{}_{E123^{*}}$\} vs  \{$f_{E*}$\}   
\end{center}

\vskip 0.2in

\newpage

% BibTeX users please use one of
%\bibliographystyle{plainnat}
%\bibliographystyle{spbasic}      % basic style, author-year citations
%\bibliographystyle{spmpsci}      % mathematics and physical sciences
%\bibliographystyle{spphys}       % APS-like style for physics
%\bibliography{}   % name your BibTeX data base

%\bibliography{mybib} %% mybib.bib

\end{document}